\documentclass[preprint,12pt,authoryear]{elsarticle}

\usepackage{xcolor}
\usepackage{amssymb}
\usepackage{booktabs}
\usepackage{lineno}
\usepackage[flushleft]{threeparttable}
\usepackage{subcaption}
\usepackage{multirow}

\journal{Journal}

\begin{document}

\begin{frontmatter}



\title{Optimal Sequential Decision-Making in Geosteering: A Reinforcement Learning Approach}


\author[uis]{Ressi Bonti Muhammad\corref{cor}}
\ead{ressi.b.muhammad@uis.no}
\author[norce]{Sergey Alyaev\corref{cor}}
\ead{saly@norceresearch.no}
\author[uis]{Reidar Brumer Bratvold}

\affiliation[uis]{organization={University of Stavanger},
            addressline={Kjell Arholms gate 41}, 
            city={Stavanger},
            postcode={4021}, 
            country={Norway}}

\affiliation[norce]{organization={NORCE Norwegian Research Centre},
            addressline={Nygårdsgaten 112}, 
            city={Bergen},
            postcode={5008}, 
            country={Norway}}

\cortext[cor]{Corresponding authors:}

\begin{abstract}
Trajectory adjustment decisions throughout the drilling process, called geosteering, affect subsequent choices and information gathering, thus resulting in a coupled sequential decision problem. Previous works on applying decision optimization methods in geosteering rely on greedy optimization or approximate dynamic programming (ADP). Either decision optimization method requires explicit uncertainty and objective function models, making developing decision optimization methods for complex and realistic geosteering environments challenging to impossible. We use the Deep Q-Network (DQN) method, a model-free reinforcement learning (RL) method that learns directly from the decision environment, to optimize geosteering decisions. The expensive computations for RL are handled during the offline training stage. Evaluating DQN needed for real-time decision support takes milliseconds and is faster than the traditional alternatives. Moreover, for two previously published synthetic geosteering scenarios, our results show that RL achieves high-quality outcomes comparable to the quasi-optimal ADP. Yet, the model-free nature of RL means that by replacing the training environment, we can extend it to problems where the solution to ADP is prohibitively expensive to compute. This flexibility will allow applying it to more complex environments and make hybrid versions trained with real data in the future.
\\
\end{abstract}


\begin{keyword}
Geosteering \sep Geosteering decisions \sep Sequential decision-making \sep Reinforcement learning \sep Machine learning \sep Subsurface energy resources
\end{keyword}

\end{frontmatter}


\newcommand{\todo}[1]{{\color{red} TODO: #1}}
\newcommand{\TODO}[1]{\todo{#1}}
\section{Introduction}
\label{sec:intro}
Geosteering operations involve a series of important decisions made by a geosteering team (GST) throughout the drilling process. 
As GST receives the new logging-while-drilling (LWD) data, the team is constantly engaged in data interpretation and utilization to make informed decisions. 
There are numerous studies about supported and automated LWD data inversion and interpretation for geosteering for both classical shallow logs \citep{alyaev2022sequential,mitkus2023bayesian,Veettil20} and deep electromagnetic measurements \citep{antonsen2022what,wang2019fast,xu2018novel,shahriari2021error}.
The automated support of decision-making given the interpretation is comparatively less studied and is the focus of our study.

The dynamic nature of geosteering operations formally establishes them as sequential decision problems \citep{Kullawan2016, KULLAWAN201890, ALYAEV2019, ALYAEV2021}. 
In recent years, several studies have proposed decision optimization methods specifically tailored for geosteering. For example, \citet{Chen2014} introduced a proactive (greedy) geosteering workflow that continuously updates the geological model with the ensemble Kalman filter, allowing rapid adaptation to new data. 
Similarly, \citet{Kullawan2014-2} developed a Bayesian, multi-objective optimization framework for geosteering, balancing diverse objectives under uncertainty. 
Although both methods enable quick decision-making, their reliance on greedy optimization can lead to suboptimal outcomes over the long term, potentially limiting their effectiveness in sequential and dynamic geosteering scenarios.

Another approach by \citet{KULLAWAN201890} uses approximate dynamic programming (ADP) method that leverages a Bayesian framework to optimize directional changes. 
While ADP provides a structured, long-term decision-making framework, its discretization involves trade-offs between accuracy and computational efficiency, which can constrain its effectiveness in capturing fine-grained dynamics in geosteering. 
Similarly, \citet{ALYAEV2019} combined the ensemble Kalman filter with ADP method, supporting multi-target geosteering through sequential model updates. 
However, these ADP-based methods lack adaptability; they are tailored to specific problem setups, making reusability across different geosteering scenarios challenging without detailed adjustments.
Lastly, \citet{Kristoffersen_2021} introduced an automatic well planner that adjusts well trajectories based on near-well geological properties, with additional optimization techniques used to refine these trajectories across various geological scenarios. However, relying on both local trajectory adjustments and supplementary optimization algorithms adds complexity, which may limit its suitability for real-time geosteering applications.

In addition to the specific disadvantages noted for each method, most of these decision support methods also require a complete model of the environment. 
This model represents all possible scenarios and the likely outcomes associated with each decision, effectively detailing how each decision impacts future conditions.
Constructing a comprehensive and precise model can be particularly difficult or even infeasible in geosteering, where subsurface conditions are often only partially understood and can change unpredictably during drilling.

This range of limitations motivates the exploration of alternative methods with greater adaptability, particularly those suited for real-time geosteering operations. 
Reinforcement learning (RL) offers a promising approach by enabling decision-making agents to develop optimal policies through iterative interactions with a simulated environment, thus eliminating the need for a predefined model \citep{sutton2018}. 
In RL, the agent refines its policy by exploring various actions and observing their outcomes, building a strategy based on direct environmental feedback rather than relying on a fixed model. 
This interaction-driven learning process allows RL to function independently of specific environment configurations, as it continuously adapts its policy based on observed transitions and rewards. 
As a result, RL demonstrates inherent flexibility across a range of applications \citep{Mnih2015-mf, Silver.2016, Gu17robot, sallab17driving} without requiring scenario-specific adjustments to the algorithm, making it particularly suitable for complex and variable geosteering conditions.

This inherent flexibility sets RL apart from the previously developed methods. 
Unlike greedy optimization, which prioritize immediate gains, RL optimizes for cumulative rewards, enabling decisions that balance both immediate and long-term objectives. 
Furthermore, RL does not depend on discretized models or complex optimization frameworks, which enhances its adaptability across diverse geosteering scenarios. 
Unlike ADP-based methods or multi-step optimization approaches, RL provides seamless transitions between environments with minimal reconfiguration. 
Its model-free design, which eliminates the need for a complete environment model, is thus crucial for addressing the varying levels of complexity and uncertainty inherent in geosteering scenarios.

In oil and gas fields, RL has mainly been used for strategic decision-making problems, in which the problems have a fairly long time window in finding the decisions. 
For instance, in waterflooding optimization, \citet{miftakhov20water} showed how RL can enhance net present value by dynamically adjusting water injection rates, achieving more stable production and lower watercut values compared to traditional methods. 
\citet{Dixit2022} demonstrated RL’s effectiveness in stochastic optimal well control, where RL was applied to determine valve openings that maximize production while managing uncertainties. 
Similarly, \citet{nasir22} highlighted RL’s adaptability by comparing it to traditional closed-loop reservoir management for optimizing bottom-hole pressures in existing wells. 
\citet{He22} proposed an RL framework for field development optimization, demonstrating that RL can effectively generalize across new scenarios. 
RL has also shown promise in operational applications requiring shorter decision intervals. For example, \citet{esp24} applied physics-informed RL to optimize electrical submersible pumps by dynamically adjusting motor frequencies. 
This approach enables real-time responses to changing operational conditions, improving production rates and energy efficiency compared to conventional constant-frequency methods.

While RL has shown potential in strategic decision-making, particularly for long-term optimizations in field development and well control, its application in operational tasks, such as geosteering, is still under explored and requires further validation. 
In this work, we aim to address this gap by applying RL within the contexts of two published geosteering examples: \citet{Kullawan2014-2} and \citet{KULLAWAN201890}. 
This approach enables a direct comparison with the original optimization methods proposed in each example—greedy optimization in the first and ADP in the second.
Both examples are relatively simple yet well-studied setups, where the information gathered while geosteering is limited solely to the interpreted distance to reservoir boundaries, without incorporating other log measurements. 
This setup provides a fair baseline for comparison with RL, enabling us to validate its suitability for tackling more complex and realistic geosteering scenarios in future work.


The following sections build upon the introduction. Section \ref{sdm} provides an overview of the Markov Decision Process (MDP) and the Bellman equation, along with detailed explanations of the three decision optimization methods used in this study: greedy optimization, ADP, and RL. 
Section \ref{RLingeos} outlines the initialization, training, and evaluation framework of the RL model, as well as the comparison method used against the other decision optimization methods. 
Section \ref{exs} presents two examples to compare and measure the performance of RL against greedy optimization and ADP methods. 
Finally, Section \ref{conc} concludes the study, summarizing the results and their implications.




\section{Sequential Decision-Making}\label{sdm}
In this section, we provide an overview of sequential decision-making, beginning with the foundational framework of the MDP and its importance in structuring these types of problems. 
We then examine various methods and algorithms developed to optimize sequential decision-making, focusing on how they tackle complex challenges and improve decision quality in dynamic environments, with particular emphasis on applications in geosteering.

The MDP is a framework for modeling decision-making problems in discrete, stochastic, and sequential domains \citep{PUTERMAN1990331}. It consists of five elements: (1) \( T \), a set of decision time points; (2) \( S \), a finite state space representing all possible configurations or conditions of the environment that the agent may encounter; (3) \( A \), a finite action space defining the set of all possible actions the agent can choose; (4) \( P \), transition probabilities that specify the likelihood of moving from one state to another given a particular action; and (5) \( R \), a real-valued reward function that provides feedback by assigning rewards or penalties based on the outcomes of the agent’s actions. 

A fundamental assumption of MDPs is the Markov property, which asserts that state transitions depend only on the current state and action, making them independent of prior states and actions.
The Markov property enables the Bellman equation, which expresses a recursive relationship between a state's value function and its successor states. \citet{sutton2018} describes this relationship as follows:
\begin{equation}\label{eq:state-value}
    V_{\pi}(s) = \sum_a \pi(a|s) \cdot \sum_{s',r}p(s',r|s,a)[r+\gamma V_{\pi}(s')],
\end{equation}
where \( V_{\pi}(s) \) represents the expected value of being in state \( s \) under policy \( \pi \). This equation divides the value into two components: the immediate reward \( r \) and the discounted value of the successor state, \( \gamma V_{\pi}(s') \). The discount factor \( \gamma \), between 0 and 1, balances the importance of immediate rewards versus future rewards, while \( \pi(a|s) \) is the probability of taking action \( a \) in state \( s \). The transition probabilities \( p(s', r|s, a) \) denote the likelihood of moving from state \( s \) to \( s' \) and receiving reward \( r \) when taking action \( a \).

Additionally, \citet{sutton2018} introduced the action-value (Q-value) function, \( Q_{\pi}(s,a) \), which represents the expected cumulative reward from taking action \( a \) in state \( s \) under policy \( \pi \):
\begin{equation}\label{eq:action-value}
    Q_{\pi}(s,a) = \sum_{s',r}p(s',r|s,a)[r+\gamma V_{\pi}(s')].
\end{equation}
The Q-value function helps identify the optimal action in each state by comparing the expected cumulative rewards for all possible actions.

In the following subsections, we explore three key decision optimization methods relevant to this study: greedy optimization \citep{Kullawan2014-2}, ADP \citep{KULLAWAN201890}, and RL. 
Each approach involves a unique formulation of the action-value function, providing distinct policies for optimizing geosteering decisions.
By examining these methods, we highlight their differences and evaluate their relative effectiveness in addressing the challenges of sequential decision-making in dynamic environments.

\subsection{Greedy Optimization}
Greedy optimization is a decision-making policy that relies solely on current knowledge and does not consider any learning of the potential impact of present decisions on future outcomes \citep{Powell2009-vj}. In the context of the Bellman equation, greedy optimization sets the discount factor $\gamma$ to 0 and we can reformulate Equation \ref{eq:action-value} as:
\begin{equation}\label{eq:myopic}
    Q(s,a) = \sum_{s',r}p(s',r|s,a)\cdot r
\end{equation} 
where the Q-value function only considers the immediate reward $r$ obtained by taking action $a$ in state $s$. 

\citet{Kullawan2014-2} proposed a decision-driven method that integrates the Bayesian framework to address multi-criteria geosteering decisions. The framework updated uncertainties ahead of the sensor location based on real-time measurements of distance to boundaries and considered only those relevant to the current decision-making stage to make decisions. Equation \ref{eq:myopic} presents the value function used by the method which does not take into account future learning and decision-making, and generally leads to locally optimal choices \citep{Kullawan2016}.

\subsection{Approximate Dynamic Programming}
Dynamic Programming (DP) is an optimization technique that aims to find optimal solutions for sequential decision problems by considering a full sequence of decisions and information, as proposed by \cite{Bellman1966}. This enables DP to evaluate the long-term consequences of each alternative and select the globally optimal decision at each decision stage. As a result, DP is more computationally demanding than the greedy optimization. In DP, the full form of Equation \ref{eq:action-value} is used.

DP and the Bellman equation have been successfully applied to sequential decision problems across various fields \citep{DP15, DP18, DP19}. However, applying DP to geosteering is challenging due to its dependence on detailed transition probabilities $p(s',r|s,a)$ \citep{sutton2018}. 
In geosteering, these probabilities require an accurate model of the subsurface environment to predict how steering actions will influence the well’s trajectory and surrounding geological conditions. 
This reliance on complete probabilistic knowledge can limit DP’s effectiveness in complex, uncertain scenarios, where constructing such detailed models is often infeasible.

To address this issue, \citet{KULLAWAN201890} proposed an ADP called the discretized stochastic dynanmic programming (DSDP) method, which approximates transition probabilities by discretizing the state space and using Monte Carlo sampling to assign state-to-state transition probabilities. 
The DSDP offers an alternative that can produce a quasi-optimal solution, which we define as the optimal solution for the discretized version of the problem, to DP with a significant reduction in computational cost.
However, applying DSDP to approximate transition probabilities across multiple geosteering scenarios may be inefficient, as each scenario necessitates algorithm adjustments, resulting in increased time and effort.

Moreover, DP and its variants are particularly susceptible to the "curse of dimensionality," where computational demands increase exponentially with the number of decision variables, alternatives, and uncertainties. 
This complexity limits the practical application of DP in real-time geosteering scenarios that require rapid decision-making. 
Consequently, this highlights the need for more adaptable decision optimization methods that can effectively manage uncertainty and reduce computational costs, all while yielding outcomes comparable to quasi-optimal solutions.

\subsection{Reinforcement Learning}\label{RLbasic}
RL refers to an optimization method for understanding and automating sequential decision-making \citep{sutton2018}. It involves a decision-making agent that learns an optimal policy in an unpredictable and complex environment. The agent interacts with the environment in an MDP system by receiving state $s_t$ and reward $r_t$ at each time step $t$ before performing action $a_t$ accordingly. The environment then transitions to a new state $s_{t+1}$ and emits a reward $r_{t+1}$ (time step increases after every interaction with the environment) in response to the action taken. The agent learns from experience and continuously adapts its policy to maximize the cumulative reward. An extended flowchart illustrating this interaction and its adaptation to our specific context is provided in Figure \ref{fig:mainloop} and described in detail in Section \ref{RLingeos}.

\subsubsection{Q-learning}
This work focuses on model-free RL, which can be further divided into action-value-based and policy-optimization-based methods \citep{sutton2018}. We work with the Q-learning method, an action-value-based method that uses an equation known as the Q-value update rule to iteratively update the Q-value based on the agent's interaction with the environment.

The Q-value update rule is as follows:
\begin{equation}\label{eq:q-learning}
    Q(s_t,a_t) \leftarrow Q(s_t,a_t) + \alpha[r_{t+1} + \gamma max_{a'}Q(s_{t+1},a') - Q(s_t,a_t)].
\end{equation}
Here, $Q(s_t,a_t)$ represents the current Q-value of performing action $a_t$ in state $s_t$. The Q-value is updated based on the immediate reward $r_{t+1}$, the maximum Q-value among all possible actions in the next state $max_{a'}Q(s_{t+1},a')$, and the current Q-value. 

The learning rate $\alpha$ controls the step size of the update, while the discount factor $\gamma$ determines the importance of the next state value. Like the DSDP, Q-learning allows for a more flexible discount factor value, which can be adjusted between 0 and 1, depending on the currently faced context.

The main difference between Q-learning, greedy optimization, and DSDP lies in the information they use to calculate the Q-values. The Bellman equation (Equation \ref{eq:action-value}) assumes that the decision-making agent has complete knowledge of the environment dynamics, or the transition probabilities $p(s',r|s,a)$, and uses this information to calculate the Q-values. On the other hand, Q-learning does not require complete knowledge of the environment dynamics, and as a result, Equation \ref{eq:q-learning} does not include the term for transition probabilities. Instead, Q-learning relies on a sequence of experiences to update the Q-values.

\subsubsection{Deep Q-Network}
Q-learning continuously updates all Q-values until convergence \citep{sutton2018}, which may not be practical for large-scale problems, especially those with continuous state spaces. The deep Q-network (DQN) was developed to overcome this limitation. DQN approximates Q-values using a deep neural network, the Q-network, which is trained to minimize a loss function that penalizes the difference between the predicted and target Q-values. 

The loss function in the DQN implementation is described as follows:
\begin{equation}\label{eq:dqn}
    L(\theta) = [y_t - Q(s_t,a_t;\theta_{t})]^2
\end{equation}
where $y_t$ is the target Q-value and $Q(s_t,a_t;\theta_{t})$ is the predicted Q-value. The Q-network is trained using stochastic gradient descent to update the weights of the network $\theta_{t}$ at each time-step $t$, ultimately leading to a better approximation of the Q-values. \cite{Mnih2015-mf} demonstrates the effectiveness of DQN for solving complex, high-dimensional RL problems by showing that the DQN agent can outperform professional humans on a range of Atari games.

Incorporating a deep neural network into a RL environment introduces the possibility of unstable or divergent learning \citep{TDdivergence}. It can be attributed to two primary causes, the correlations between sequences of experiences, $e_t = (s_t,a_t,r_{t+1},s_{t+1})$, that may lead to highly correlated data distribution and the correlations between action-value and target-value \citep{Mnih2015-mf}.

To address the first cause, DQN uses experience replay, which involves storing the experience at each time step in a memory data set, $D_t = \{e_1,\dots,e_t\}$, and uniformly sampling a minibatch of experience $b \sim U(D)$ from the memory data set to update the Q-network weights using the samples. The second issue is addressed by introducing a separate network, the target network $\hat{Q}$, with weight parameters $\theta^-$. While the initial network weight parameter $\theta$ is updated at each iteration, the target network weight parameter $\theta^-$ is updated every $C$ iteration step by cloning the weight parameters of the initial network. Thus, the target-value in Equation \ref{eq:dqn} becomes:
\begin{equation}\label{eq:target}
y_t = r_{t+1}+\gamma max_{a'}\hat{Q}(s_{t+1},a';\theta_{t}^-).
\end{equation}



DQN is currently limited to solving problems with a discrete action space consisting of a finite set of possible actions. Each action is typically assigned a unique identifier (e.g., "one," "two") or index (e.g., "0," "1"). This differs from the continuous action space, where an action is selected from a certain distribution. To tackle problems with continuous action spaces, another branch of RL called the policy optimization-based method is needed.

To apply DQN to geosteering decision-making contexts, it is necessary to ensure that the action space is discrete. While this limitation may seem restrictive, it does not diminish the value of DQN compared to greedy optimization and DSDP, which are likewise confined to discrete action space problems. On the other hand, DQN offers an advantage over DSDP due to its more straightforward implementation. The RL agent (or DQN agent) is independent of its environment (see Figure \ref{fig:mainloop}), enabling the same agent to be trained across multiple contexts without altering the method.

Table \ref{tab:algotable} summarizes the differences between the sequential decision-making optimization methods described in this study.

\begin{table}[h]
\centering
\begin{threeparttable}
\caption{Comparison of sequential decision-making optimization methods}
\label{tab:algotable}
\begin{tabular}{|c|c|c|c|c|}
\hline
\textbf{Algorithms} &
  \textbf{Type} &
  \textbf{\begin{tabular}[c]{@{}c@{}}Future \\ Information\end{tabular}} &
  \textbf{\begin{tabular}[c]{@{}c@{}}Action \\ Space\end{tabular}} &
  \textbf{\begin{tabular}[c]{@{}c@{}}State \\ Space\end{tabular}} \\ \hline\hline
\begin{tabular}[c]{@{}c@{}}\textbf{Greedy}\\ \citep{Kullawan2014-2}\end{tabular}  & Model-based & \begin{tabular}[c]{@{}c@{}}Not\\ Considered\end{tabular}   & Discrete & Discrete                                                          \\ \hline
\begin{tabular}[c]{@{}c@{}}\textbf{DSDP}\\ \citep{KULLAWAN201890}\end{tabular}     & Model-based & \begin{tabular}[c]{@{}c@{}}Fully\\ Considered\end{tabular} & Discrete & Discrete                                                          \\ \hline
\textbf{Q-learning} & Model-free  & \begin{tabular}[c]{@{}c@{}}Implicitly\\ Considered\tnote{*}\end{tabular}       & Discrete & Discrete \\ \hline

\textbf{DQN} & Model-free  & \begin{tabular}[c]{@{}c@{}}Implicitly\\ Considered\tnote{*}\end{tabular}       & Discrete & \begin{tabular}[c]{@{}c@{}}Discrete and\\ Continuous\end{tabular} \\ \hline
\end{tabular}
\begin{tablenotes}
\footnotesize
\item[*] In this context, 'implicitly considered' means that Q-learning and DQN estimate future information or state values through observed interactions, not through a direct model of the environment.
\end{tablenotes}
\end{threeparttable}
\end{table}

\section{Reinforcement Learning in Geosteering}\label{RLingeos}
In this section, we outline the methodology used in this study, detailing the training and evaluation processes for the RL agent applied to geosteering scenarios. 
We begin with the initialization of the RL agent and the geosteering environment, which includes defining the state and action spaces along with other relevant information. 
Next, we describe the general training and evaluation procedures, emphasizing how the agent interacts with diverse geological conditions to enhance its decision-making capabilities. 
Finally, we provide an overview of the DQN architecture used in this study.

\subsection{General Training and Evaluation}\label{gentrain}
\begin{figure}[ht]
    \centering
    \includegraphics[width=1\textwidth]{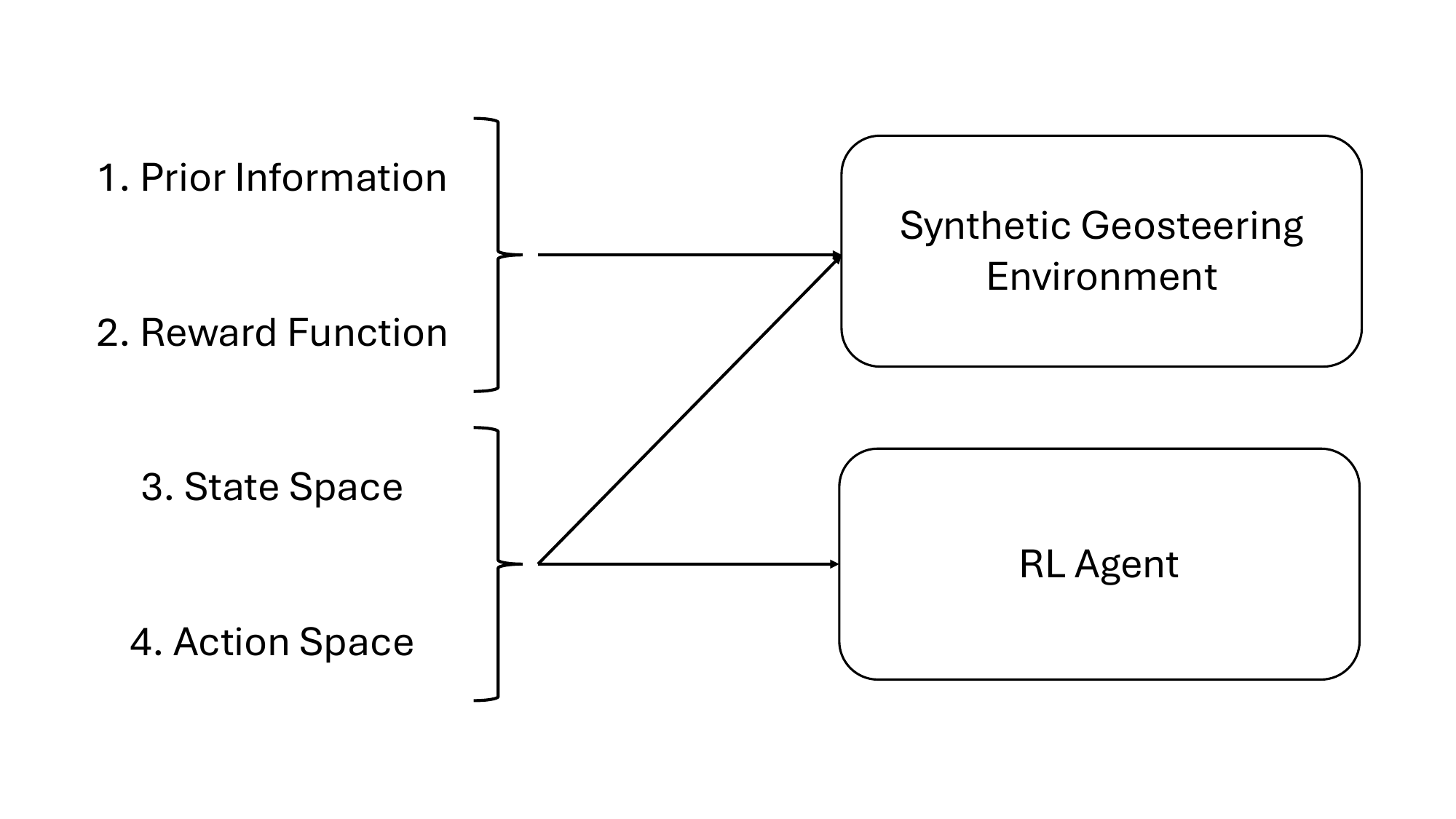}
    \caption{Key components required for initializing the RL agent and geosteering environment before training. This setup includes defining state and action spaces, establishing the reward function, and incorporating prior geological data} 
    \label{fig:init}
\end{figure}

Before training begins, several initialization steps are required to configure the RL agent and the geosteering environment, as shown in Figure \ref{fig:init}. 
These steps include specifying the state and action spaces to define the scope of the agent’s interactions within the environment, as well as establishing the reward function, which guides the agent’s learning through feedback. 
Additionally, prior information, such as relevant geological data and constraints, is incorporated to set initial conditions for each training iteration, commonly referred to as an episode in RL.

Here, we assume that the geosteering environment is equipped with the necessary equations and logic to update the well trajectory based on the agent’s selected actions. 
For instance, it may use the minimum curvature method or similar techniques to calculate trajectory changes, ensuring that the state updates accurately reflect the selected actions.
In more general terms, this means the environment contains predefined, deterministic rules that define how the state transitions occur in response to each action. 
With these components in place, the agent is ready to interact with the environment iteratively.

\begin{figure}[ht]
    \centering
    \includegraphics[width=0.99\textwidth]{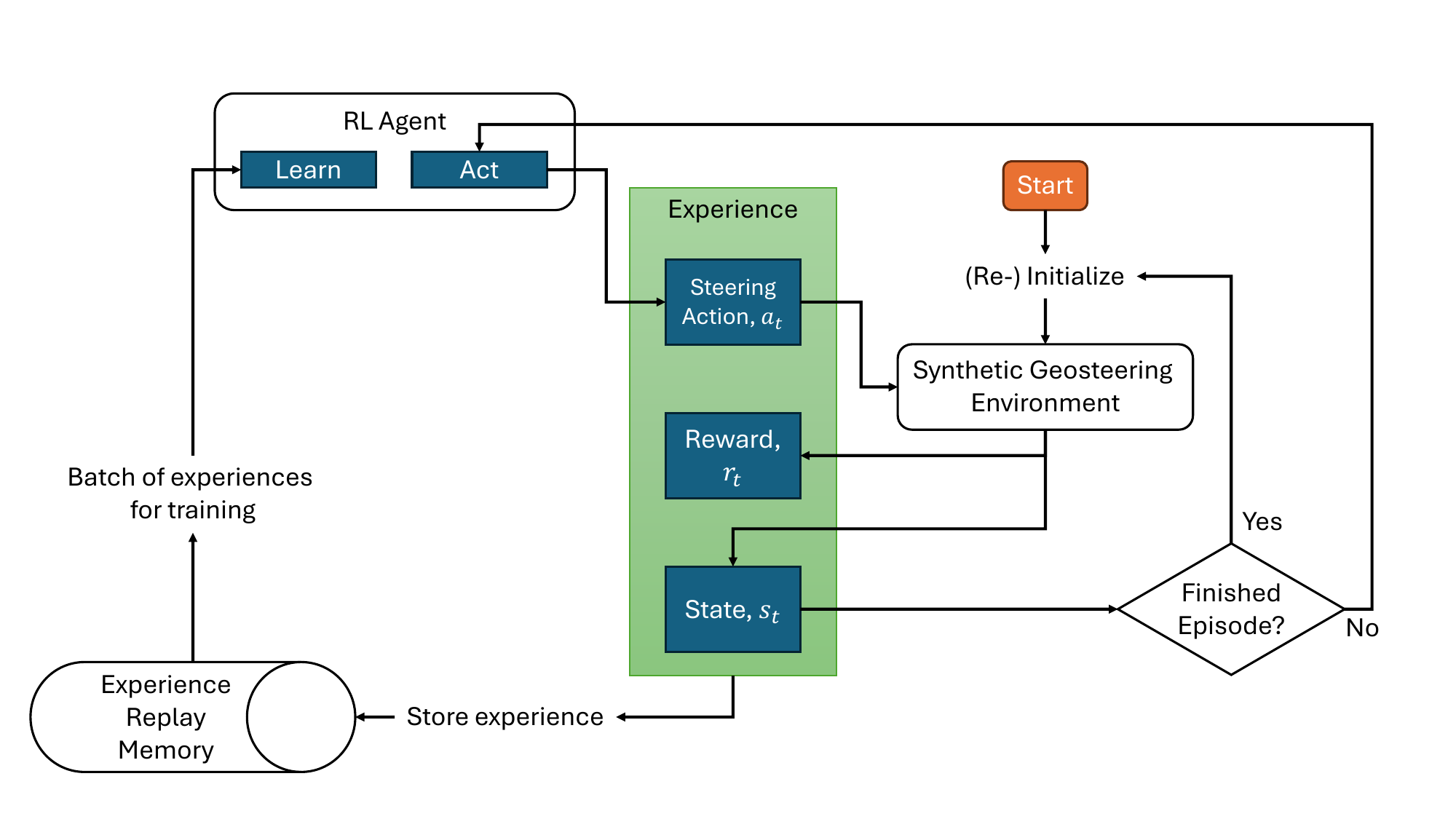}
    \caption{Extended RL training flowchart, building on the standard agent-environment interaction. Uncertain variables introduce variability at the start of each episode, simulating diverse geological conditions. Experience replay memory stores interactions for batch training, enhancing stability and generalization across scenarios.} 
    \label{fig:mainloop}
\end{figure}

Building on the standard RL interaction framework, where an agent interacts with an environment by observing states, taking actions, and receiving rewards (see subsection \ref{RLbasic}), the extended flowchart in Figure \ref{fig:mainloop} incorporates additional elements relevant to our context. 
This setup includes an initialization process at the beginning of each new episode, which reconfigures the environment to simulate diverse geological conditions. 
These variations may include differences in target layer thickness, fault quantity, and fault displacement magnitude. 
By ensuring that each episode presents unique scenarios, this approach is crucial for training the agent to handle a range of geosteering challenges, enhancing its adaptability to realistic conditions.

The flowchart also highlights the role of experience replay memory, a technique commonly used in off-policy RL algorithms like DQN. 
Off-policy algorithms enable the agent to learn from past experiences that were not necessarily collected while following the current policy, allowing for more efficient use of data and supporting stable training.

During each episode, the agent stores its experiences—state, action, reward, and next state—in a memory buffer, rather than updating its policy immediately after each interaction. 
Periodically, the agent samples a randomized batch from this replay memory for training.
Using this batch, the agent updates the loss function and target value equations (refer to Equations \ref{eq:dqn} and \ref{eq:target}), optimizing its policy based on varied past interactions.
By learning from this randomized batch, the agent mitigates correlations between consecutive experiences, which stabilizes the learning process and helps it generalize across different scenarios.

We train different RL agents tailored to each specific problem setup derived from the variations in geological models mentioned in \citet{Kullawan2014-2} and \citet{KULLAWAN201890}. This approach ensures that the decision-making policy of each agent is finely tuned to the unique characteristics and challenges presented by the different examples. The training samples are generated differently for each example and will be explained in their respective subsections. Additionally, to ensure an unbiased comparison with greedy optimization and DSDP, we train the RL agent using 51 different random seeds for each setup.

We evaluate the performance of each trained agent by generating 1000 geological realizations from the same distributions used during training. We then compute the median of the average reward obtained by the 51 trained agents. We name this median as "RL-Robust." RL-Robust is calculated as follows:
\begin{equation}\label{eq:median}
    r_{RL} = \mathrm{Median}\left(\bar{r}_1,\bar{r}_2,\ldots,\bar{r}_{m}\right)
\end{equation}
\begin{equation}\label{eq:mean}
\bar{r} = \frac{\sum_{i=1}^{n} r_{i}}{n} 
\end{equation}
where $m$ equals to 51 and $\bar{r}$ is the average reward out of $n = 1000$ geological realizations. 
Note that Equation \ref{eq:median}, or the RL-Robust method, is exclusively used by RL methods due to the training of 51 different agents. In contrast, greedy optimization and DSDP use a simple average calculation from 1000 geological realizations as they do not involve training decision-making agents.

All computations are carried out on a system with an 11th Gen Intel(R) Core(TM) i5-1135G7 CPU $@$ 2.4GHz (8 CPUs) and 16GB of RAM, with no GPU acceleration used for RL training or evaluation. 
The software environment is built in Python, with libraries essential for RL and data analysis. 
The RL code primarily uses PyTorch \citep{pytorch} for the DQN implementation and NumPy \citep{harris2020array} for additional data calculations and manipulation. 
On the other hand, the greedy and DSDP methods are implemented mainly with NumPy for all data calculations and manipulations. 
Matplotlib \citep{hunter07} is used across all methods for plotting results.

\subsection{Model Architecture}
\begin{figure}[ht]
    \centering
    \includegraphics[width=0.9\textwidth]{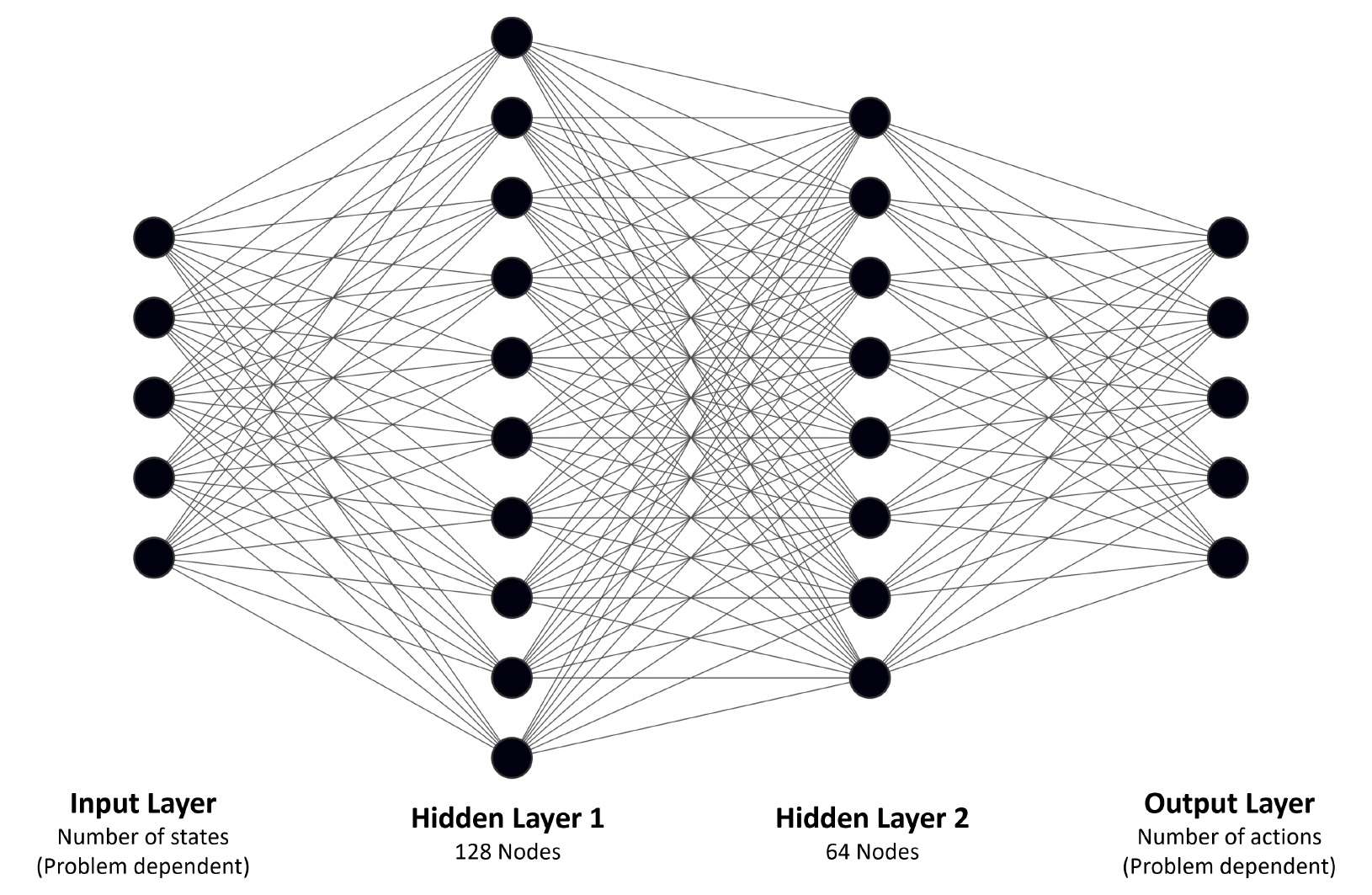}
    \caption{General neural network architecture used in the study, generated using \citet{LeNail2019}. The neural network architecture consists of two hidden layers that take the state space as inputs and output the Q-values for each available action. The input and output layer configurations, which vary based on the state and action spaces specific to each environment, are detailed in Table \ref{tab:RL1state} and last paragraph of subsection \ref{rlset2}.}
    \label{fig:arch}
\end{figure}

The model (or the RL agent in Figure \ref{fig:init} and \ref{fig:mainloop}) uses a deep neural network to estimate the Q-value for a given state-action pair. 
Logically, the network takes the state-action pair as input and produces the approximate Q-value as output. 
However, the $\epsilon$-greedy policy used in the method requires comparing the Q-values of all available actions in a state, resulting in a linear increase in cost as the number of possible actions increases. 
To address this issue, \citet{Mnih2015-mf} proposes an alternate network design that uses the state space as input and outputs the Q-value for each action, allowing for a single forward pass to estimate the Q-value of a given state.
Figure \ref{fig:arch} illustrates the general network architecture used in the study.

In this study, we use PyTorch to define our neural network architecture and train the model. 
We choose a relatively simple two-hidden-layer deep neural network instead of fully using the proposed design, which uses convolutional hidden layers to capture images as its input. 
Our network consists of two hidden fully-connected linear layers with a ReLU activation function, with the first and second hidden layers containing 128 and 64 rectifier units, respectively. 
The input and output layers are configured according to the state and action spaces specific to each environment. 
For instance, in the first example, the input layer corresponds to the state dimensions \{$DTUB_i$, \dots, $DTUB_{i+n}$, \dots\}, while in the second example, it aligns with a different set of inputs tailored to that scenario (see Table \ref{tab:RL1state} and last paragraph of subsection \ref{rlset2} for detailed definitions of the input and output layers in each example).
While the number of input and output nodes varies based on these dimensions, the core DQN structure (number of hidden layers and nodes) remains unchanged.
This setup allows the model to be easily applied across different geosteering scenarios without altering the main architecture.

\section{Numerical Example}\label{exs}
In this section, we show the application of RL by studying two examples from published journals. Specifically, we analyze the studies did by \citet{Kullawan2014-2} and \citet{KULLAWAN201890} and show how RL can optimize decision-making in complex and uncertain environments. We compare the results of RL to those obtained using greedy optimization and DSDP. By comparing these results, we can gain valuable insights into the abilities of RL to generate near-optimal decision guidance and objective results.

\subsection{First Example}\label{ex1}
This subsection presents the application of RL to the geosteering environment defined in \citet{Kullawan2014-2}, where the authors used greedy optimization for optimizing the geosteering decisions. The geosteering scenario involves drilling a horizontal well in a three-layered model with non-uniform reservoir thickness and quality. The reservoir consists of a sand layer sandwiched between shale layers and is divided into two permeability zones: a high-quality zone in the top 40 percent of the reservoir and a low-quality zone comprising the remaining 60 percent.

Given the non-uniform thickness of the reservoir, primary uncertainties in this geosteering example are the depths of the top and bottom reservoir boundaries. Signals are received from the sensor, which is located at or behind the bit, and these signals are used to determine the distance to the boundaries (at or behind the bit). These distances are then, in turn, used to update uncertainties about boundary locations ahead of the bit using a Bayesian framework. The framework assumes that the signals are accurate.

The reservoir boundaries are discretized into $N$ points. Steering decisions are made by adjusting the well inclination at every $n$ discretization point, where the change in inclination at each decision stage is restricted to no more than $5^o$. At each decision stage, there are 11 alternatives for inclination adjustment, ranging from $-5^o$ to $5^o$ in increments of $1^o$. The minimum curvature method is used to calculate the trajectory of the well at each discretization point based on the chosen inclination change.

\begin{figure}[ht]
    \centering
    \includegraphics[scale=0.48]{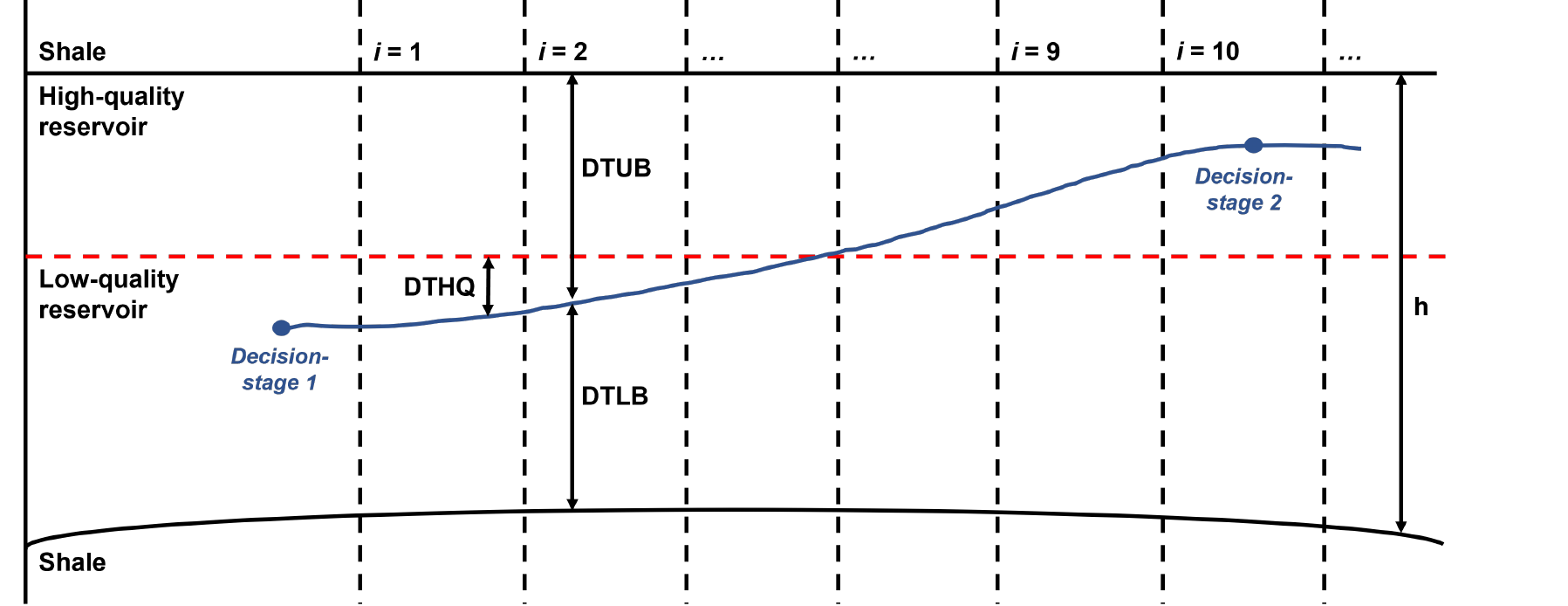}
    \caption{Illustration of the geosteering scenario, a remake based on \citet{Kullawan2014-2}. At every n = 10 discretization points, decisions are made to determine the well trajectory. The blue line represents the well path, while the red dashed line denotes the boundary between high and low-quality reservoir zones. The thickness of the reservoir at each discretization point is represented by $h$, while $DTUB$ and $DTLB$ denote the distances from the well to the upper and lower reservoir boundaries, respectively. Additionally, $DTHQ$ denotes the distances from the well to the high-quality zone.}
    \label{fig:env1}
\end{figure}

Figure \ref{fig:env1} shows the geosteering environment as described above. The blue line represents the well trajectory drilled in the three-layered model. The red dashed line indicates the boundary between the high-quality and low-quality reservoir zones. Decisions are made at every $n=10$ discretization point to determine the path of the well. The thickness of the reservoir varies at each discretization point, denoted by $h$. The distances from the well to the upper and lower reservoir boundaries are denoted by $DTUB$ and $DTLB$, respectively, so $h = DTUB + DTLB$. In addition, $DTHQ$ denotes the distance from the well to the high-quality zone.

The primary objective of the example is to to maximize the length of the well in the high-quality reservoir zone with specific penalty for reservoir exits. It is implemented by the two objective functions from \citet{Kullawan2014-2}.
The first reward function, $r_{1,i}$ at discretization point $i$, is: 
\begin{equation}\label{eq:reward1}
    r_{1,i} = 14.654x_i^3 - 17.778x_i^2 + 7.2252x_i
\end{equation}
where $x_i$ is the distance between the well and the nearest reservoir boundary, min($DTUB, DTLB$), normalized by the reservoir thickness $h_i$. The function takes its maximum value (1) when the well is placed in the middle of the reservoir ($x_i = 0.5$). The second objective, $r_{2,i}$ at discretization point $i$, is:
\begin{equation}\label{eq:reward2}
    r_{2,i} = -2\times10^{-5}y_i^2 + 0.009y_i.
\end{equation}
Here, $y_i$ represents the permeability of the zone, and the equation is constructed for a scenario where the maximum permeability is 200 mD, which results in a value of 1 for $y_i = 200$.

The decision context described by Equations \ref{eq:reward1} and \ref{eq:reward2} is multi-objective with two conflicting objectives. Equation \ref{eq:reward1} indicates that placing the well in the center of the reservoir minimizes the risk of exiting the reservoir, while Equation \ref{eq:reward2} suggests that the well should be positioned in the high-quality zone to maximize its value. However, positioning the well in the high-quality zone increases the risk of drilling into the upper shale layer, which goes against the primary objective.

We thus apply a common multi-objective \citep{bratvold2010making} decision analytic approach where the objectives are weighted. This entails computing the weighted overall reward for each of the 11 alternatives. The weighted overall reward, $r_j$, at decision stage $j$, is:
\begin{equation}\label{eq:RLreward}
r_{j} = w_1\sum_{i = n * (j-1)}^{n * j - 1}r_{1,i} + w_2\sum_{i = n * (j-1)}^{n * j - 1}r_{2,i}.
\end{equation}
Here, $w_1$ and $w_2$ represents the weights assigned to the two objectives and $w_1 + w_2 = 1$. The computed reward for the $i^{th}$ discretization point for the two objectives are represented by $r_{1,i}$ and $r_{2,i}$, respectively.
Given that $n$, the number of discretization points between decisions, is set to 10 for this context, the maximum value of $r_j$ is therefore 10. 
This maximum is achieved under the condition that $r_{1,i}$ and $r_{2,i}$ each reach their maximum value of 1 at all 10 discretization points in the interval.

\subsubsection{RL Setting}\label{rlset1}
In order to use RL method for sequential decision-making, we need to define parameters such as the action space, reward function, and state space based on the geosteering scenario. These parameters are needed for the RL agent to learn and find the optimal policy for the scenario.

\textbf{Action space.} The geosteerers are engaged in decision-making to place the well optimally. They will make decisions, each with a set of alternatives, consisting of 11 discrete values ranging from -$5^o$ to $5^o$ at each decision stage. As a result, the neural network has 11 output nodes in this scenario, corresponding to the available actions that the RL agent can take.

\textbf{Reward function.} One of the objectives in \citet{Kullawan2014-2} was to study the effect of alternative weights for each objective on the final well placement rather than the reward function. However, the RL agent is trained to maximize the reward function defined in Equation \ref{eq:RLreward} for every available scenario. We compare greedy optimization and RL based on their reward function to ensure consistent training and evaluation. We use $N=100$ discretization points, and have 10 decision stages. Thus, the maximum reward for a single geosteering equals to $n * 10 = 100$.

\textbf{State space.} We need to specify a state space relevant to the reward function to ensure that the RL agent has the necessary information to optimize geosteering decisions. Specifically, we provide the RL agent with $n+1$ pieces of information for reservoir thickness and the vertical distances to the reservoir boundaries and high-quality zone, with the $n$ pieces representing posterior updates from the Bayesian framework, ahead of the sensor location, and the $+1$ piece representing the sensor reading at the current decision stage. In addition, the RL agent receives another 5 pieces of information. These include the horizontal distance from the starting point, the inclination assigned to the current discretization point, the quality of the reservoir, and the weights associated with each objective. With $n = 10$ or 10 points between each decision stage, the RL agent receives 49 state information inputs, consisting of 40 posterior information pieces, 4 sensor readings, and the 5 complementary information pieces mentioned above. This configuration is referred to as "RL-Posterior."

In an alternative method, the Bayesian framework and posterior updates are not used. Instead, the RL agent receives distance-to-boundaries sensor readings from $n$ discretization points behind the sensor location, while other complementary information is identical in both methods. This method can help us show the capacity of RL (DQN) to optimize the problem by implicitly predicting the boundaries ahead of the sensor location without the need for potentially costly Bayesian computations. This configuration is referred to as "RL-Sensor."

The state and action spaces of both RL methods are summarized in Table \ref{tab:RL1state}. Notably, the action space is identical for both methods. The first two rows of the state space correspond to the $n+1$ information described earlier. However, the important distinction lies in the subscript notation of the information. For the RL-Posterior method, the information range from the sensor location $i$ to $n$ discretization points ahead, while for the RL-Sensor method, the information starts from $n$ discretization points behind the sensor location $i$. The remaining information, including $Inc_i$ representing the inclination at the current point, $i$ denoting the current discretization point (or equivalently, the horizontal distance from the starting point), $y$ representing the reservoir quality, and $w_1$ and $w_2$ denoting the weights of each objective, remain consistent across both agents.

\begin{table}[ht]
\centering
\caption{Comparison of the state space (input) and action space (output) of the two RL methods for the first example.}
\label{tab:RL1state}
\begin{tabular}{|c|c|c|}
\hline
\textbf{Model} & \textbf{State Space} & \textbf{Action Space} \\ \hline
\textbf{RL-Posterior} &
  \begin{tabular}[c]{@{}c@{}}\{$DTUB_i$,\dots, $DTUB_{i+n}$,\\ $DTLB_i$,\dots, $DTLB_{i+n}$,\\ $DTHQ_i$,\dots, $DTHQ_{i+n}$,\\ $h_i$,\dots, $h_{i+n}$,\\ $Inc_i$, $i$, $y$, $w_1$, $w_2$\}\end{tabular} & \multirow{2}{*}{\begin{tabular}[c]{@{}c@{}}\rule{0pt}{2.6ex}\{-$5^o$, -$4^o$, \dots, $4^o$, $5^o$\}\end{tabular}} \\ \cline{1-2}
\textbf{RL-Sensor} &
  \begin{tabular}[c]{@{}c@{}}\{$DTUB_{i-n}$,\dots, $DTUB_{i}$,\\ $DTLB_{i-n}$,\dots, $DTLB_{i}$,\\ $DTHQ_{i-n}$,\dots, $DTHQ_{i}$,\\ $h_{i-n}$,\dots, $h_{i}$,\\ $Inc_i$, $i$, $y$, $w_1$, $w_2$\}\end{tabular} & \\ \hline
\end{tabular}
\end{table}

\subsubsection{Training Results}
In the first example, we face the challenge of working with only a single geological realization from \citet{Kullawan2014-2}. To address this, we develop a simple forward function that mirrors the characteristics of the provided realization. This forward function is then used to generate the training and evaluation samples.

Following \citet{Kullawan2014-2}, we split the example into two scenarios based on the reservoir quality disparity between the two zones. We train the RL agent to optimize both scenarios simultaneously to avoid the need to train the RL agent multiple times. It could also show the generalizability of the resulting decision-making policy.

\begin{figure}[ht]
  \centering

  \begin{subfigure}{0.5\textwidth}
    \centering
    \includegraphics[width=\linewidth]{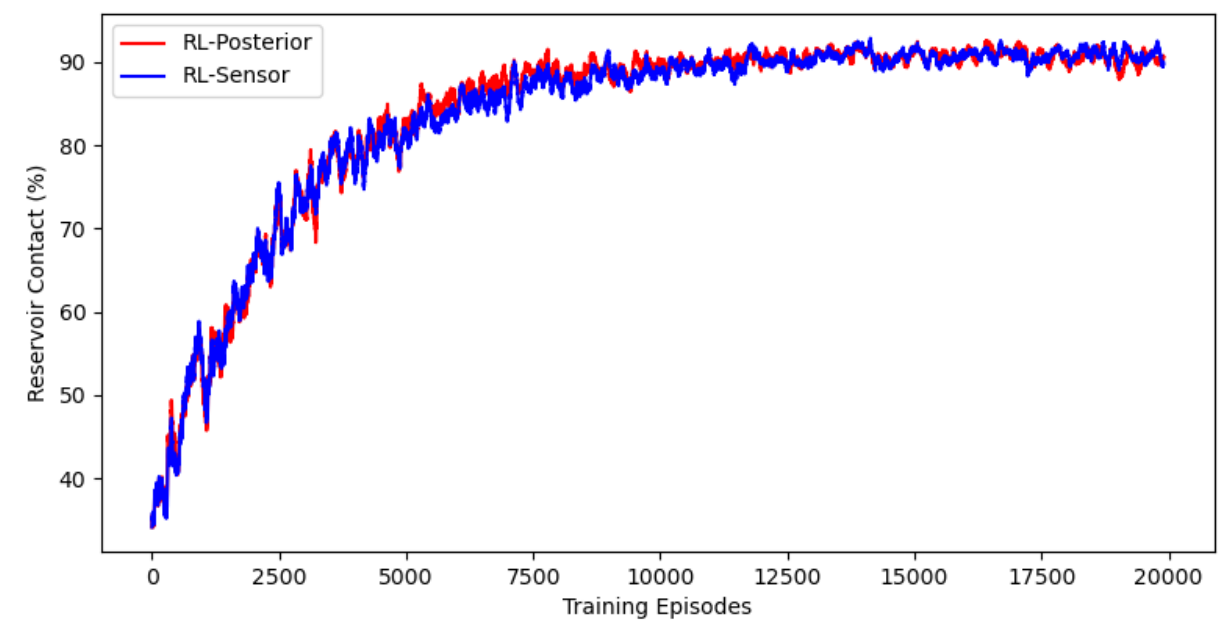}
    \caption{Reservoir Contact}
    \label{fig:RL1-rc}
  \end{subfigure}%
  \hfill
  \begin{subfigure}{0.5\textwidth}
    \centering
    \includegraphics[width=\linewidth]{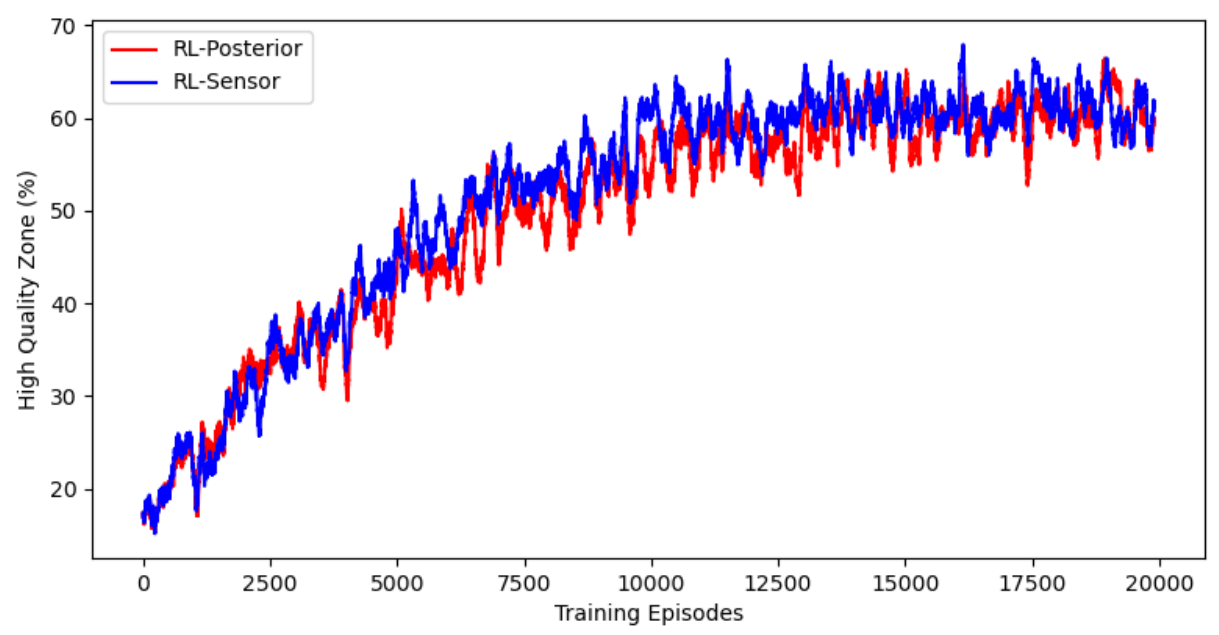}
    \caption{High Quality Zone}
    \label{fig:RL1-hiq}
  \end{subfigure}%
  \hfill
  \begin{subfigure}{0.5\textwidth}
    \centering
    \includegraphics[width=\linewidth]{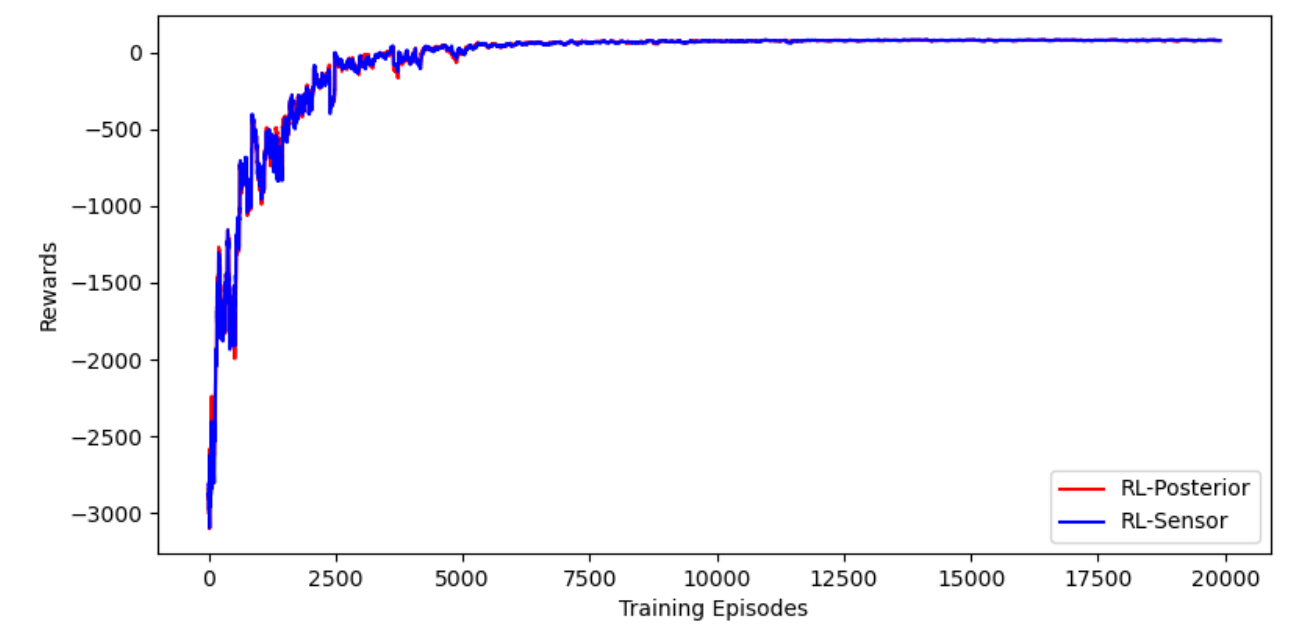}
    \caption{Rewards}
    \label{fig:RL1-rewards}
  \end{subfigure}

  \caption{Evolution of individual objectives (reservoir contact and high-quality zone percentages) and the overall rewards of two RL agents during training. The red lines represent the RL-Posterior method, while the blue lines represent the RL-Sensor method. The figure reflects the average from the last 100 training episodes.}
  \label{fig:RL1train}
\end{figure}

Figure \ref{fig:RL1train} visually shows the outcomes of training a single seed of the two RL methods: the RL-Posterior and RL-Sensor, in the first example. The red lines represent the RL-Posterior method, while the blue lines represent the RL-Sensor method. The figure presents the evolution of the percentage of each individual objective, specifically the reservoir contact and the high-quality zone percentages, alongside the rewards obtained during the geosteering operation. The figure is constructed by taking the average of the numbers (percentages and rewards) obtained during the last 100 training episodes. This approach provides a smoothed representation of the performance of an RL decision-making agent and helps mitigate the effects of short-term fluctuations. 

Both RL agents gradually improve their decision-making policies throughout the training, as seen by the overall increase in all objectives and rewards. The reservoir contact objective, which initially stands at less than 40 percent during the early stage of training, improves significantly for both agents, reaching approximately 90 percent by the end of the training. Additionally, both agents increase the high-quality zone percentage from 20 percent to approximately 60 percent after the training sequence. 

Regarding the rewards, it is important to address the presence of large negative values observed at the beginning of the training. These values are caused by the reward function associated with the first objective. The reward function assigns a large negative number when the well trajectory is significantly distant from the target zone, resulting in the noticeable negative values shown in the figure. However, as the RL agent learns from experience, the rewards gradually increase and remain above zero when the training sequence ends.

While the difference between the two agents may appear negligible at first glance, closer examination reveals a slight advantage for the RL-Sensor method regarding the high-quality zone objective. It is important to note that the figure shown is based on a single training seed, and a more comprehensive comparison will be done in the subsequent section. In that section, the performance of all training seeds will be averaged, providing a more robust comparison of the relative performance between the RL agents.

Another important aspect is the computational cost of training a single seed. The RL-Posterior method requires approximately 2500 seconds to complete the training sequence, whereas the RL-Sensor method completes the training in a significantly shorter time, around 500 seconds. This substantial difference in training duration is caused by the utilization of the Bayesian framework in the RL-Posterior method, which introduces additional computational costs. On the other hand, the RL-Sensor method, which does not rely on the Bayesian framework, offers a more computationally efficient alternative.

\subsubsection{Evaluation Results}
\begin{figure}[ht]
  \centering
  
  \begin{subfigure}[b]{\textwidth}
    \includegraphics[width=\linewidth]{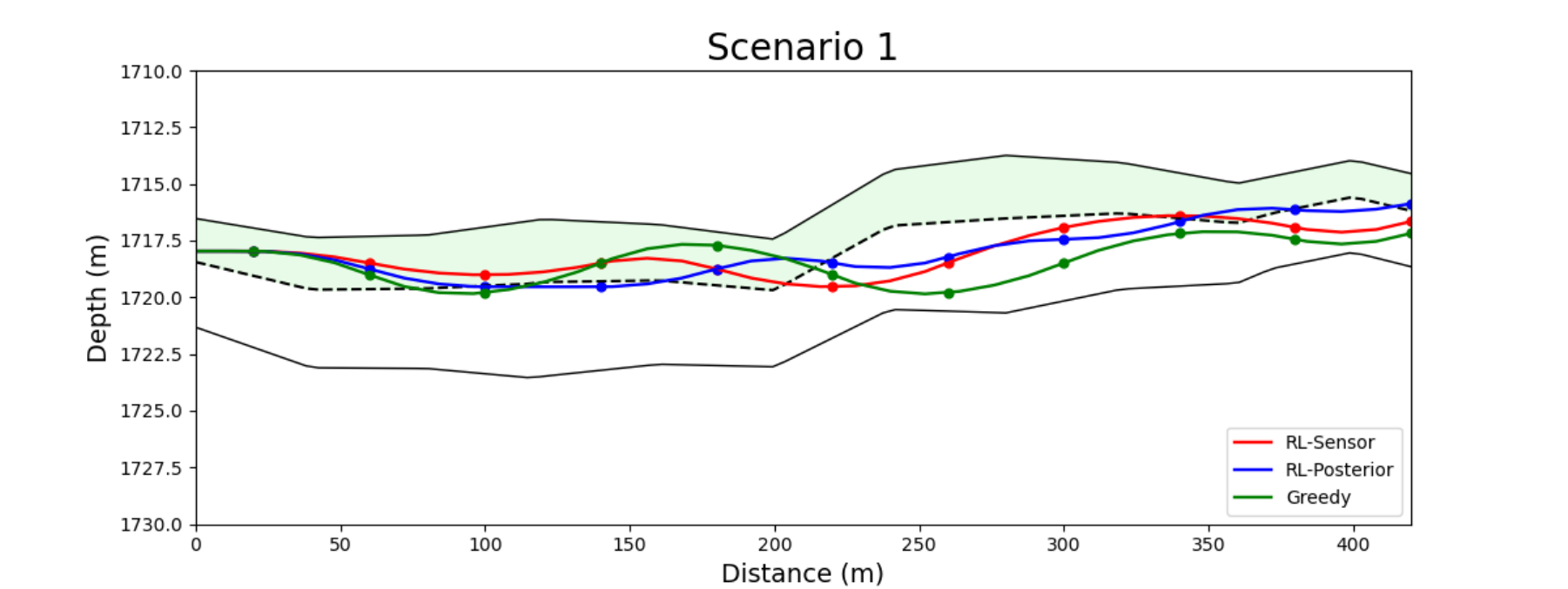}
  \end{subfigure}
  
  \begin{subfigure}[b]{\textwidth}
    \includegraphics[width=\linewidth]{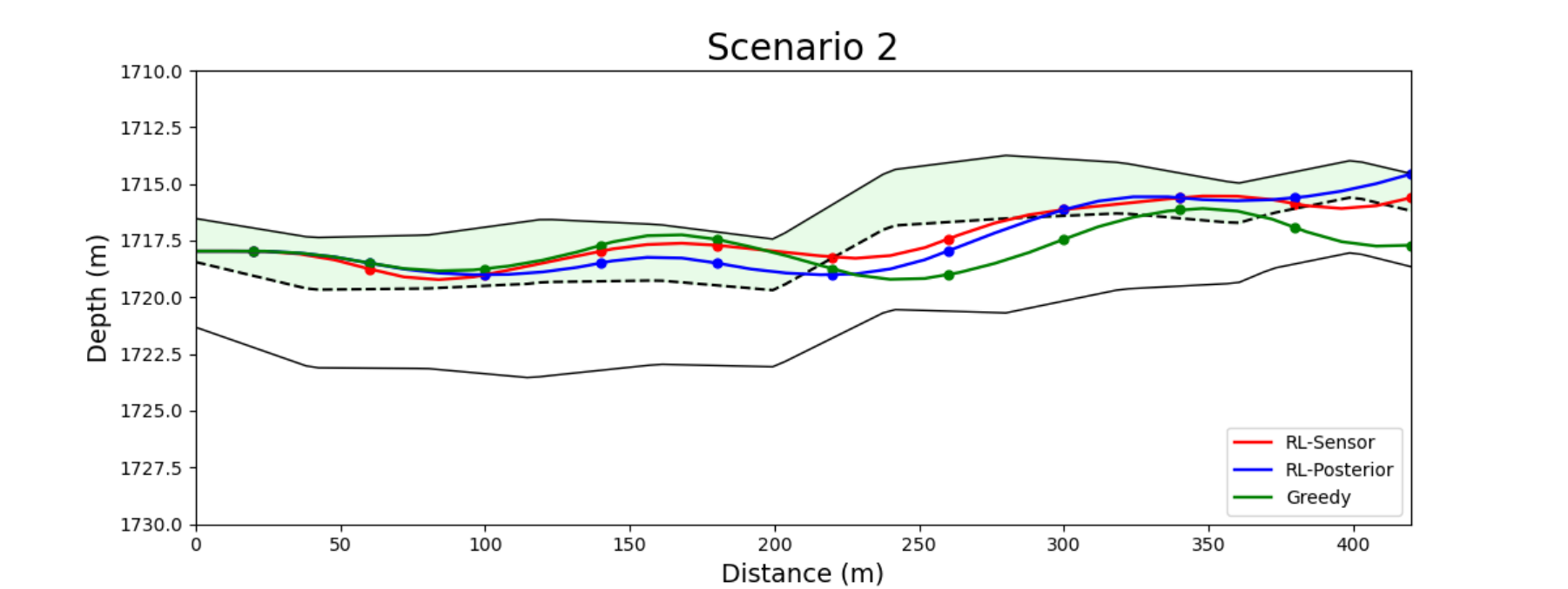}
  \end{subfigure}
  
  \caption{Comparison of resulting geosteering trajectories from greedy optimization, RL-Posterior, and RL-Sensor methods for a single geological realization from the first example. The dashed line marks the separation of high and lower quality zones, with the green shaded areas indicating the high-quality zones.}
  \label{fig:ex1single}
\end{figure}

Figure \ref{fig:ex1single} shows well trajectories for a single geological realization using greedy optimization, RL-Posterior, and RL-Sensor, shown in various colors. The solid black lines represent the reservoir boundaries, and the dashed line separates the high from lower quality zones, with the green shaded areas indicating the high-quality zones.

Additionally, this subsection presents the results for the evaluation explained in subsection \ref{gentrain}, while the subsequent subsection will focus on discussing those results in more detail.

\textbf{Scenario 1.} The permeability of the high-quality zone is 200 mD, whereas the permeability of the low-quality zone is 100 mD. There is a distinction in quality, although it is not substantial. Consequently, the first objective is assigned a greater weight, with $w_1 = 0.67$ and $w_2 = 0.33$. 

The top panel of Figure \ref{fig:ex1single} presents the results for Scenario 1 from a single geological realization. Given a higher $w_1$, the RL agents are trained to prioritize maintaining reservoir contact. Both the RL-Posterior and RL-Sensor agents achieve 100 percent contact with the reservoir. Similarly, greedy optimization manages to maintain all segments of the well trajectory inside the reservoir. However, some of its segments are steered close to the reservoir boundaries, unlike the trajectories from both RL agents, which are more centrally aligned.

\begin{table}[ht]
\centering
\caption{Scenario 1 - Average results for greedy optimization and Median results for both RL-agents,}
\begin{tabular}{|c|r|r|r|}
\hline
\textbf{Methods}  & \multicolumn{1}{c|}{\textbf{Rewards}}   & \multicolumn{1}{c|}{\textbf{\begin{tabular}[c]{@{}c@{}}Reservoir\\ Contact(\%)\end{tabular}}}  & \multicolumn{1}{c|}{\textbf{\begin{tabular}[c]{@{}c@{}}High\\ Quality(\%)\end{tabular}}}    \\ \hline\hline
\begin{tabular}[c]{@{}c@{}}Greedy\\ \citep{Kullawan2014-2}\end{tabular}           & 71.44                                   & 87.90                                                 & 43.00                                             \\ \hline
RL-Posterior     & 85.25                                   & 92.86                                                 & 44.06                                             \\ \hline
RL-Sensor       & 85.79                                   & 92.86                                                 & 46.77                                             \\ \hline
\end{tabular}
\label{tab:sce1-100}
\end{table}

Table \ref{tab:sce1-100} presents the evaluation results of all three methods. The median reward of the RL-Posterior method is 85.25 out of 100, representing a significant improvement of 19.34 percent over the greedy optimization. Meanwhile, the RL-Sensor method achieves a slightly higher median reward of 85.79. Furthermore, both RL agents yield better well placement by achieving higher figures on both objectives. The median values of RL in this and all subsequent tables refer to the RL-Robust calculation method, as detailed in Subsection \ref{gentrain}. This method is applied consistently throughout the study.

\textbf{Scenario 2.} The permeability of the high-quality zone is 200 mD, while the permeability of the low-quality zone is 20 mD. Given the more substantial difference in quality compared to scenario 1, it is important to position the well in the top portion of the reservoir. Therefore, the weight for the second objective is greater than in the first objective, with $w_1 = 0.41$ and $w_2 = 0.59$.

The bottom panel of Figure \ref{fig:ex1single} illustrates the outcomes for Scenario 2 in a single realization. With higher $w_2$, the RL agents prioritize steering the well into the high-quality zone. Both RL agents, RL-Posterior and RL-Sensor, succeed in staying within the high-quality zone for the majority of the geosteering operation, achieving about 79 percent coverage. In contrast, despite maintaining its reservoir contact to 100 percent, greedy optimization struggles to steer into the high-quality zone, particularly in the later segments of the well trajectory, resulting in only 59 percent of its trajectory being within the high-quality zone.

\begin{table}[ht]
\centering
\caption{Scenario 2 - Average results for greedy optimization and Median results for both RL-agents.}
\begin{tabular}{|c|r|r|r|}
\hline
\textbf{Methods}    & \multicolumn{1}{c|}{\textbf{Rewards}} & \multicolumn{1}{c|}{\textbf{\begin{tabular}[c]{@{}c@{}}Reservoir\\ Contact(\%)\end{tabular}}}  & \multicolumn{1}{c|}{\textbf{\begin{tabular}[c]{@{}c@{}}High\\ Quality(\%)\end{tabular}}}   \\ \hline\hline
\begin{tabular}[c]{@{}c@{}}Greedy\\ \citep{Kullawan2014-2}\end{tabular}              & 55.57                                 & 82.50                                                 & 53.20                                             \\ \hline
RL-Posterior       & 73.76                                 & 89.85                                                 & 65.62                                             \\ \hline
RL-Sensor          & 74.21                                 & 88.65                                                 & 70.22                                             \\ \hline
\end{tabular}
\label{tab:sce2-100}
\end{table}

Table \ref{tab:sce2-100} presents the average results of greedy optimization and median results of both RL configurations in scenario 2. Similar to scenario 1, the RL-Posterior and RL-Sensor methods outperform greedy optimization in all three metrics, with higher rewards, reservoir contact percentage, and high-quality percentage. 

The primary objective in this scenario is to position the well inside the high-quality zone. As a result, the well is placed closer to the upper reservoir borders, which increases the probability of exiting the reservoir, leading to a relatively low reward from the first reward function. Furthermore, the low-quality zone in scenario 2 has a lower permeability value than in scenario 1, reducing the overall reward of scenario 2. Nevertheless, the table illustrates that RL has higher value in scenario 2 than in scenario 1, with a 32.74 percent increase for the RL-Posterior method and a 33.55 percent increase for the RL-Sensor method. 

\subsubsection{Discussion}
The RL methods used in this study yield significantly better results than greedy optimization. The superior performance of the RL methods is attributed to their ability to learn from experience and explore different policies to maximize the reward. By contrast, greedy optimization only considers the immediate reward and does not consider future decisions and future learning.

The results also show that the RL-Sensor method consistently outperforms the RL-Posterior method in terms of median reward across all scenarios. Although the difference in performance is insignificant, it is consistent, and the RL-Sensor method achieves the results without requiring posterior updates, suggesting that it can implicitly anticipate reservoir boundaries using the sensor readings behind the sensor location. One plausible explanation for the observed difference is that the posterior updates derived from the Bayesian framework might contain inaccuracies and errors. Although the implicit predictions from RL can also suffer from similar issues, the superior performance of the RL-Sensor method implies that the implicit predictions provide more accurate approximations compared to the posterior updates. 

Moreover, eliminating the Bayesian framework in the RL-Sensor method reduces the training time for a single seed from 2500 to 500 seconds while still achieving superior results. With a computational cost that is 5 times cheaper, the RL-Sensor method remains the superior method even when its results are slightly worse than those of the RL-Posterior method. Therefore, we exclusively use the RL-Sensor method for the second example described in the following subsection. 

\subsection{Second Example}\label{ex2}
In this section, we show the application of the RL-Sensor method to an example previously shown by \citet{KULLAWAN201890}. Their study used greedy optimization from \citet{Kullawan2014-2} and the proposed DSDP to optimize a geosteering scenario in thin and faulted reservoirs. The geosteering context for this example is similar to the first example, where a horizontal well is drilled in a three-layered reservoir model consisting of a sand reservoir sandwiched between shale layers.

The example considers a constant reservoir thickness and uniform quality, reducing uncertainties in this case to the depths of the upper reservoir boundary and the location and displacement of faults. The prior knowledge of the depth of boundaries is combined with fault information, such as the number of faults, expected fault displacement, and possible fault location. As in the first example, \citet{KULLAWAN201890} uses a Bayesian framework to update the combined prior information based on real-time sensor data. Additionally, the study assumes that the real-time information is accurate.

\begin{figure}[ht]
    \centering
    \includegraphics[scale=0.62]{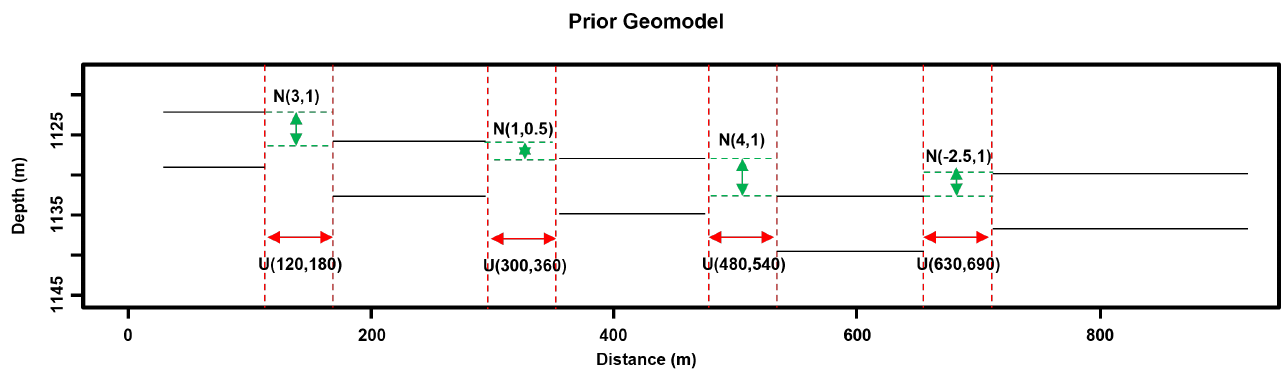}
    \caption{Illustration of prior geomodel and fault uncertainties remade based on \citet{KULLAWAN201890}. The expected upper and lower boundaries are shown as solid black lines. Uncertainty in potential fault displacements is represented by green dashed lines, modeled using a normal distribution. Uncertainty in possible fault locations is represented by red dashed lines, modeled using discrete uniform distributions.}
    \label{fig:priormodel}
\end{figure}

We use the same prior geomodel and fault uncertainty parameters as \citet{KULLAWAN201890}, shown in Figure \ref{fig:priormodel}. The reservoir boundaries are discretized into 30 points ($N=30$) spaced 30 meters apart, with solid black lines indicating the expected upper and lower boundaries of the reservoir. Green dashed lines represent potential fault displacements, with uncertainty modeled using a normal distribution. Red dashed lines indicate possible fault locations, with uncertainty modeled using discrete uniform distributions. For instance, the first fault has an estimated displacement of 3 meters, with a standard deviation of 1 meter, and may be located 120 meters, 150 meters, or 180 meters from the first discretization point.

In this geosteering setting, there are 29 decision stages, and steering decisions are made at every discretization point ($n = 1$) by directly modifying the well depth without using the minimum curvature method. At each decision stage $j$, 5 options are available for altering the well depth, ranging from $-0.5$-m TVD to $0.5$-m TVD with a $0.25$-m TVD increase. Additionally, sidetracking can be done if the well exits the reservoir after deciding to modify or maintain the well depth. We refer to it as the default setup, where a steering decision is followed by a sidetrack decision. Sidetracking can ensure that the well is drilled towards the center of the boundary. However, it incurs additional costs. In other words, sidetracking represents a trade-off between future gain and present increased expense.

\begin{figure}[ht]
    \centering
    \includegraphics[scale=0.75]{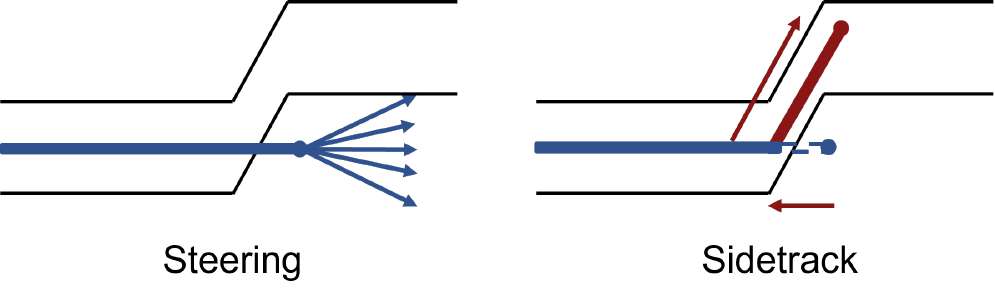}
    \caption{Illustration of alternatives from the second example. The blue line represents the well trajectory, while the black lines represent the reservoir boundaries. The decision maker can make a steering decision using the 5 alternatives represented by the blue arrows. If the sidetrack is chosen, the well is taken back to the previous decision stage and drilled directly to the middle of the reservoir boundaries for the next decision stage, as shown by the red line and arrows.}
    \label{fig:actionex2}
\end{figure}

Figure \ref{fig:actionex2} illustrates how each alternative changes the well depth at each decision node. In this example, the blue line represents the well trajectory, which has exited the reservoir boundaries, represented by the black lines. When making a steering decision, the decision maker can adjust the well depth using any of the 5 alternatives represented by the blue arrows. On the other hand, if the decision maker chooses to execute a sidetrack, the well is taken back to the previous decision stage and drilled directly to the middle of the reservoir boundaries for the next decision stage, as shown by the red line and arrows. The blue dashed line in the figure indicates that the previous well trajectory is discarded as if it had never been drilled.

In this example, we aim to optimize the reward of a geosteering project by maximizing reservoir contact and minimizing operating costs. To this end, the setting uses a reward function $r_j$ that combines the values $v_j$ and operating costs $c_j$ at each decision stage $j$. Specifically, $r_j$ is defined as:

\begin{equation}\label{eq:reward_ex2}
r_{j} = v_j - c_j.
\end{equation}
Here, $v_j$ represents the value given the location of the well, which is equal to the production value $v_{prod}$ if the well is located within the reservoir and 0 otherwise. The operating cost $c_j$ is the sum of the drilling cost $c_d$, and the sidetrack cost $c_{ST}$ if sidetracking is done. We use values of $v_{prod}$ ranging from 0.5 to 4.0, with $c_d$ and $c_{ST}$ held constant at 0.0625 and 2.567, respectively. Consequently, for 29 decision stages, the maximum value of a geosteering project is $29\cdot v_{prod}$, and the minimum operating cost is $29\cdot c_d = 1.81$.

\subsubsection{RL Setting}\label{rlset2}
As in the first example, we need to specify the following three parameters from the geosteering scenario to use RL for sequential decision-making:

\textbf{Action space.} In the default setup, there are two decision nodes in one decision stage: the first for modifying the drill bit depth and the second for performing a sidetrack. However, this configuration can lead to sub-optimal results for specific scenarios when we use RL. For example, when the value $v_{prod}$ is less than the sidetrack cost $c_{ST}$, an RL decision-making agent using two decision nodes may follow a greedy decision-making policy and decline to do a sidetrack. Conversely, when $v_{prod}$ exceeds $c_{ST}$, the agent will always choose to sidetrack, irrespective of the actual value of the well. Thus, using two decision nodes could limit the ability of the RL agent to make optimal decisions regarding sidetracking.

To address this issue, we adapt the action space to include a single decision node with 6 alternatives, 5 of which correspond to adjustments in the drill bit depth, while the sixth alternative is for performing a sidetrack. As a result, the neural network for the second scenario contains 6 output nodes. 

However, this modification introduces another issue where the sidetrack alternative is consistently available, whereas in the default setup, the sidetrack option is restricted when the drill bit is already within the reservoir. To address this issue, we temporarily force the RL agent to disregard the sidetrack alternative when it is unnecessary. On the other hand, if the drill bit exits the reservoir, every alternative is available for the RL agent. Hence, it can either execute a sidetrack or continue making steering decisions for the subsequent decision stages. This approach enables the agent to make appropriate decisions based on the current location of the drill bit. As a result, we can evaluate the ability of the RL agent to handle both steering and sidetrack decisions effectively.

\textbf{Reward function.} The reward function used in the RL configuration remains unchanged and is the same as the default setting, as shown in Equation \ref{eq:reward_ex2}. Thus, the RL agent receives a reward $r_j$ after each decision stage. Moreover, we did not alter the evaluation method since the default setting already used a consistent approach that compared the performance of different methods based on their total reward function, which is defined as the sum of rewards over all decision stages, i.e., $\sum r_j$.   

\textbf{State space.} As the second example assumes a constant reservoir thickness and homogeneous reservoir quality, the RL-Sensor method requires less information about the reservoir than in the first. We provide an RL decision-making agent $n+1$ information on vertical distances between the well trajectory and reservoir boundaries. Additionally, with the inclusion of faults and sidetracks in the geosteering setting, we also inform the agent about the possible zone (start and end points) and the expected displacement of the subsequent fault as shown in Figure \ref{fig:priormodel}. 

Furthermore, it is important for the agent to know whether the well trajectory is inside or outside the reservoir. We also inform the agent about its horizontal distance from the initial point and the value of the current geosteering project. With $n = 1$, the RL-Sensor method gathers 10 pieces of information at each decision stage, which serve as inputs for the neural network.

The state and action spaces of the RL-Sensor for the second example are as follows: the state space of the RL-Sensor model includes sensor readings on the distance to reservoir boundaries ($DTUB$ and $DTLB$), information on impending faults (the start and end points of the faults, and the mean displacement), an exit indicator, the horizontal distance from the starting point, and a randomized production value ($v_{prod}$). The action space consists of five predefined steering decisions $\{-0.5, -0.25, 0, +0.25, +0.5\}$, with an additional option for sidetracking when necessary.

\subsubsection{Training Results}

\begin{figure}[ht]
  \centering

  \begin{subfigure}{0.5\textwidth}
    \centering
    \includegraphics[width=\linewidth]{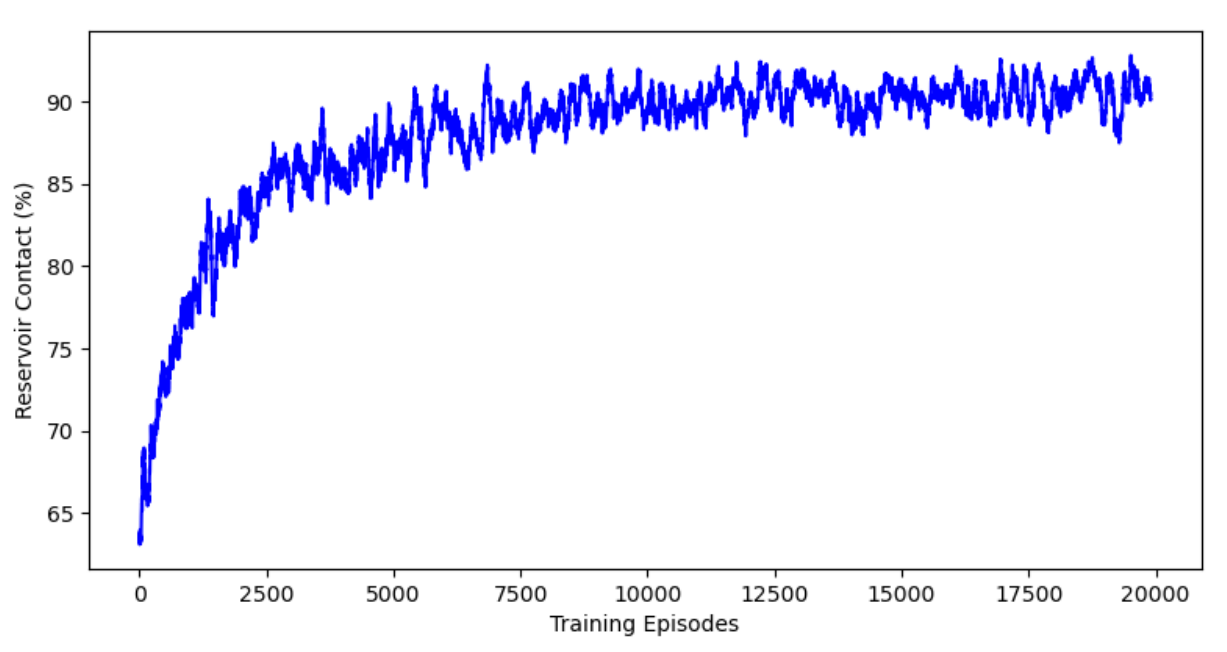}
    \caption{Reservoir Contact}
    \label{fig:RL2-rc}
  \end{subfigure}%
  \hfill
  \begin{subfigure}{0.5\textwidth}
    \centering
    \includegraphics[width=\linewidth]{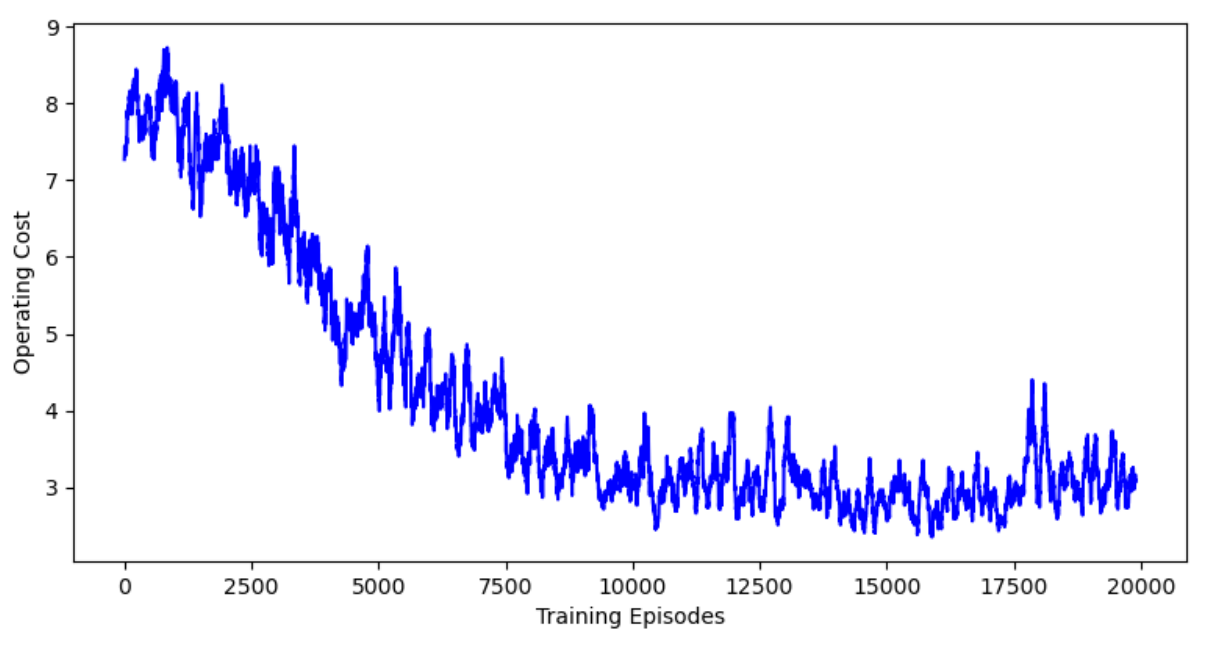}
    \caption{Operating Cost}
    \label{fig:RL2-cost}
  \end{subfigure}%
  \hfill
  \begin{subfigure}{0.5\textwidth}
    \centering
    \includegraphics[width=\linewidth]{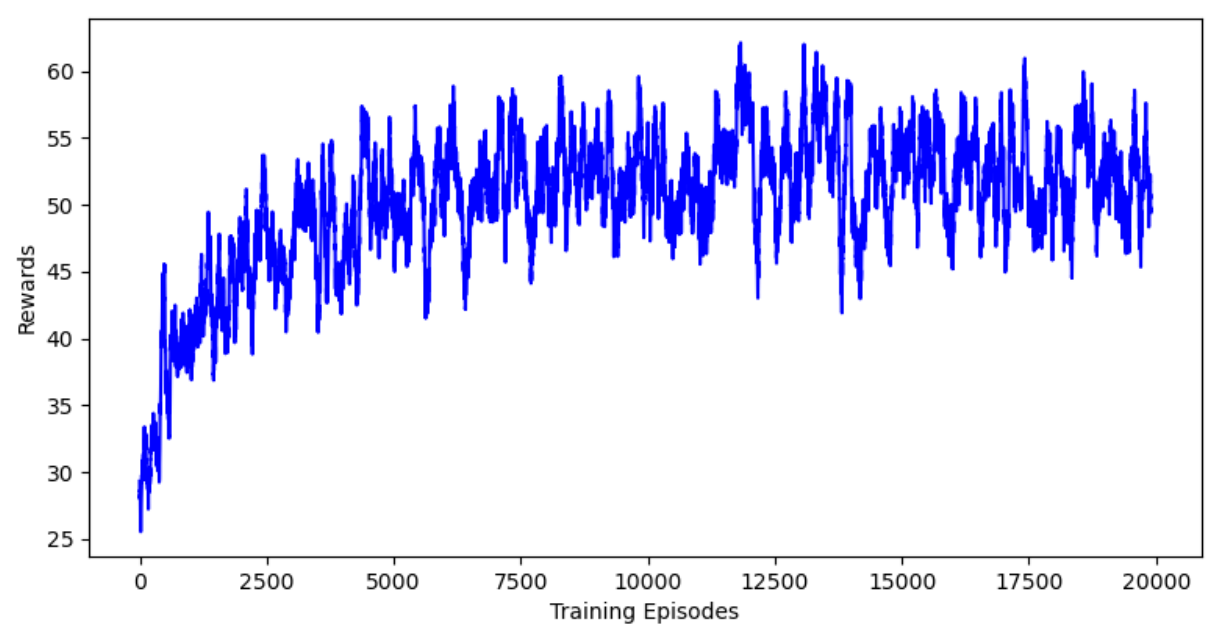}
    \caption{Rewards}
    \label{fig:RL2-rewards}
  \end{subfigure}

  \caption{Evolution of individual objectives (reservoir contact and operating costs) and the overall rewards of the RL-Sensor method during training. The figure reflects the average from the last 100 training episodes}
  \label{fig:RL2train}
\end{figure}

In the second example, prior information on the geological realization shown in Figure \ref{fig:priormodel} includes probability distributions of fault information. These distributions allow us to generate a wide array of training and evaluation samples efficiently, eliminating the need for a forward function as in the first example.

Figure \ref{fig:RL2train} illustrates the training outcomes of the RL-Sensor method in the second example. The training for a single seed requires approximately 20 minutes. Similar to the first example, the figure presents the evolution of individual objectives, specifically the reservoir contact and the operating cost, alongside the overall rewards obtained during the geosteering operation. The figure is also constructed by taking the average obtained during the last 100 training steps. Moreover, the training also considers the number of scenarios the example has. The scenario in this example depends on the production value $v_{prod}$ that ranges from 0.5 to 4.

The figure illustrates the evolution of the RL-sensor method decision-making policy. Over time, the RL-Sensor method improves the reservoir contact from an initial percentage of 65 to approximately 90 percent. Simultaneously, it reduces the operating cost from a starting cost of 8 to around 3, which correlates with decreased sidetrack operations. In other words, the RL-Sensor method gradually learns to make a better steering decision, reducing the need for frequent sidetrack operations.

The evolution of the average rewards shows a higher fluctuation level than in the previous example. This increased variability is primarily attributed to the influence of the production value, $v_{prod}$, which is determined by the random sampling procedures used during the training sequence. For instance, when $v_{prod}$ is set to 0.5, the maximum reward attainable by a decision-making agent in a single training episode is $29 \times 0.5 - 1.81 = 12.69$. Conversely, when $v_{prod}$ equals 4, the maximum achievable reward becomes $29 \times 4 - 1.81 = 114.19$, approximately 10 times larger than the other scenario.

\subsubsection{Evaluation Results}
\begin{figure}[ht]
  \centering
  
  \begin{subfigure}[b]{\textwidth}
    \includegraphics[width=\linewidth]{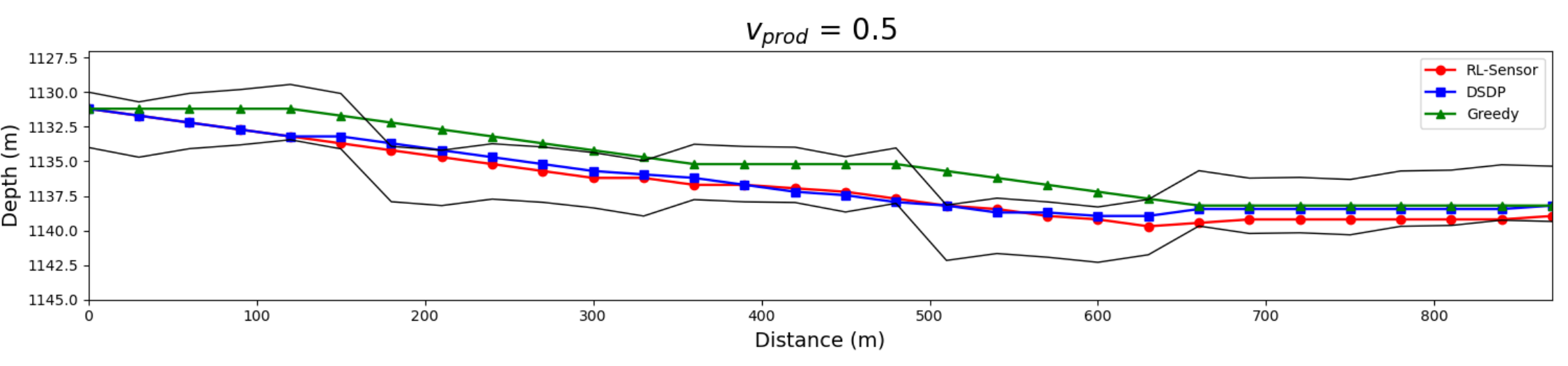}
  \end{subfigure}
  
  \begin{subfigure}[b]{\textwidth}
    \includegraphics[width=\linewidth]{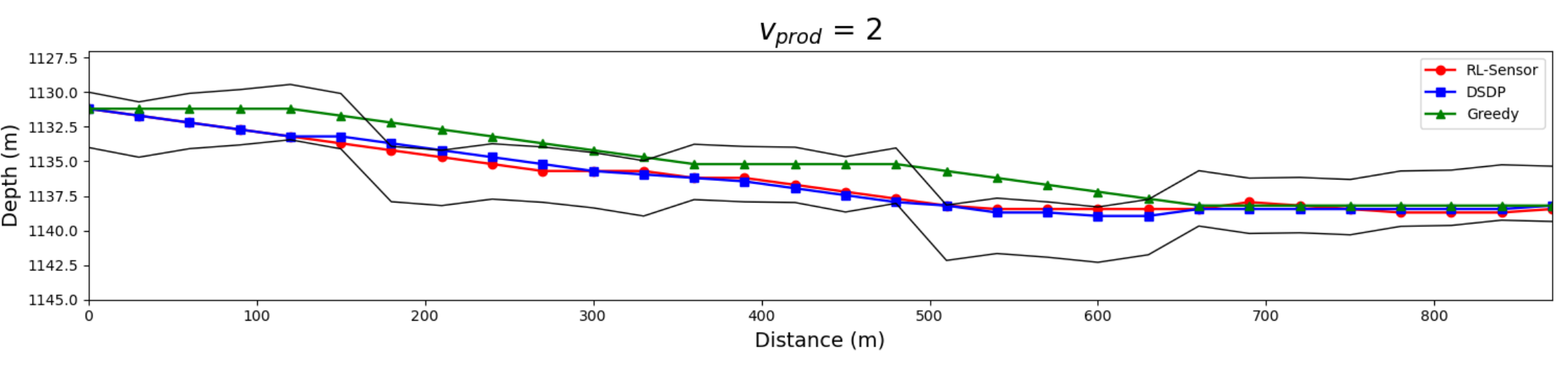}
  \end{subfigure}

  \begin{subfigure}[b]{\textwidth}
    \includegraphics[width=\linewidth]{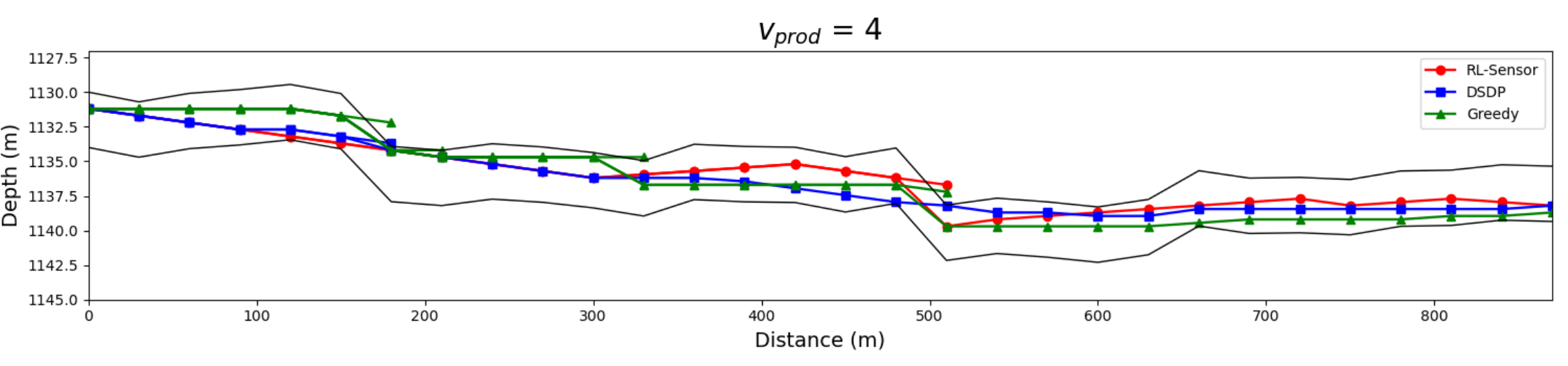}
  \end{subfigure}
  
  \caption{Comparison of resulting geosteering trajectories from greedy optimization, DSDP, and RL-Sensor methods for a single geological realization from the second example.}
  \label{fig:ex2single}
\end{figure}

Figure \ref{fig:ex2single} shows well trajectories for a single geological realization from the second example. The well trajectories, from greedy optimization, DSDP, and RL-Sensor, are shown in different colors while the solid black lines represent the reservoir boundaries. There are three panels showing the impact of varying $v_{prod}$ values on the well trajectories and overall reward functions.

Additionally, this subsection presents the results for the evaluation explained in subsection \ref{gentrain} by using the prior information shown in Figure \ref{fig:priormodel}, while the subsequent subsection will provide a discussion and analysis of the results.

\textbf{Scenario 1. }$v_{prod} = 0.5$, resulting in a very expensive sidetrack cost and making sidetrack decisions less favorable than steering decisions ($v_{prod} <<< c_{ST}$). 

In the top panel of Figure \ref{fig:ex2single}, we observe that for $v_{prod} = 0.5$, DSDP and RL-Sensor methods yield well trajectories that closely mirror each other, effectively maintaining the well trajectory within the reservoir boundaries without the need for sidetrack decisions. In contrast, the trajectory from greedy algorithm diverges from the reservoir on two occasions before steering back into it without making any sidetrack decision. Given that none of the methods incur the additional costs of sidetracking, all share an identical operating cost of 1.81. Consequently, the overall reward is determined solely by reservoir contact. Here, the RL-Sensor has a slight advantage over DSDP, which briefly exits the reservoir just before the 200-meter mark, and is significantly more efficient than greedy optimization.

\begin{table}[ht]
\centering
\caption{Scenario 1 - Results for Greedy Optimization, DSDP, and RL-Sensor.}
\begin{tabular}{|c|r|r|r|}
\hline
\textbf{Methods}     & \multicolumn{1}{c|}{\textbf{Rewards}}    & \multicolumn{1}{c|}{\textbf{\begin{tabular}[c]{@{}c@{}}Reservoir\\ Contact(\%)\end{tabular}}} & \multicolumn{1}{c|}{\textbf{\begin{tabular}[c]{@{}c@{}}Operating\\ Cost\end{tabular}}} \\ \hline\hline
\begin{tabular}[c]{@{}c@{}}Greedy\\ \citep{Kullawan2014-2}\end{tabular}            & 8.33                      & 69.94                                  & 1.81                                 \\ \hline
\begin{tabular}[c]{@{}c@{}}DSDP\\ \citep{KULLAWAN201890}\end{tabular}              & 11.40                     & 92.47                                  & 2.01                                 \\ \hline
RL-Sensor         & 11.50                     & 91.80                                  & 1.81                                 \\ \hline
\end{tabular}
\label{tab:sce1}
\end{table}

Table \ref{tab:sce1} shows the evaluation rewards, reservoir contact, and operating cost of all three methods for scenario 1. At $v_{prod} = 0.5$, greedy optimization never chooses to do a sidetrack, as suggested by its minimum operating cost of 1.81. It achieves the lowest average reservoir contact of 69.94 percent and, unsurprisingly, the lowest average reward among the other methods, at 8.33. On the other hand, the DSDP achieves a significantly higher reservoir contact of 92.47 percent compared to greedy optimization. 

The RL-Sensor method achieves a high reservoir contact of 91.80 percent while maintaining a minimal cost of operation. In this scenario, the RL-Sensor decision-making policy is similar to greedy optimization in that it does not do sidetrack operations and relies only on steering decisions. Despite not performing any sidetracks, the results show that the RL-Sensor method,  with a median reward of 11.50, slightly outperforms the DSDP. 

\textbf{Scenario 2. }$v_{prod} = 2$, which is still slightly lower than the sidetrack cost ($v_{prod} < c_{ST}$). 

The middle panel of Figure \ref{fig:ex2single} demonstrates a similarity with the finding in the top panel, where DSDP and RL-Sensor trajectories closely align, consistently steering their trajectory within the reservoir boundaries without any sidetrack decisions. There is also no changes for the trajectory from greedy optimization, as the $v_{prod}$ is still lower than the $c_{ST}$. The overall reward mirrors the outcomes of Scenario 1, with the RL-Sensor marginally outperforming DSDP and showing significantly greater efficiency over greedy optimization.

\begin{table}[ht]
\centering
\caption{Scenario 2 - Results for Greedy Optimization, DSDP, and RL-Sensor.}
\begin{tabular}{|c|r|r|r|}
\hline
\textbf{Methods}   & \multicolumn{1}{c|}{\textbf{Rewards}}      & \multicolumn{1}{c|}{\textbf{\begin{tabular}[c]{@{}c@{}}Reservoir\\ Contact(\%)\end{tabular}}} & \multicolumn{1}{c|}{\textbf{\begin{tabular}[c]{@{}c@{}}Operating\\ Cost\end{tabular}}} \\ \hline\hline
\begin{tabular}[c]{@{}c@{}}Greedy\\ \citep{Kullawan2014-2}\end{tabular}           & 38.76                     & 69.96                                  & 1.81                                 \\ \hline
\begin{tabular}[c]{@{}c@{}}DSDP\\ \citep{KULLAWAN201890}\end{tabular}             & 51.93                     & 95.02                                  & 3.18                                 \\ \hline
RL-Sensor        & 52.31                     & 96.54                                  & 3.68                                 \\ \hline
\end{tabular}
\label{tab:sce2}
\end{table}

Table \ref{tab:sce2} shows the evaluation rewards, reservoir contact, and operating cost of all three methods for scenario 2. The decision-making policy of greedy optimization remains the same at $v_{prod}$ = 2 as the production value is still below the sidetrack cost. As a result, greedy optimization still achieves the minimum operating cost with no improvement in reservoir contact. Similar to the first scenario, its average reward is the lowest among all methods at 38.76. On the other hand, the DSDP performs more sidetrack operations than in the previous scenario, resulting in a higher cost. However, this method achieves an average reward of 51.93, considerably better than greedy optimization.

Like the DSDP, the RL-Sensor method incurs a higher operating cost than in scenario 1 due to performing more sidetrack operations. Specifically, the decision-making policy of the RL-Sensor method leads to an operating cost of 3.68 while achieving a high reservoir contact of 96.54 percent. These results show the ability of the RL-Sensor method to strike a balance between maximizing reservoir contact and minimizing operating cost, even when the production value is still below the sidetrack cost. Compared to the DSDP, the RL-Sensor method yields a slightly higher median reward of 52.31. 

\textbf{Scenario 3. }$v_{prod} = 4$, resulting in a very cheap sidetrack cost and making it, when available, more favorable than steering decisions ($v_{prod} >>> c_{ST}$). 

In the bottom panel of Figure \ref{fig:ex2single}, the results differ from the earlier scenarios, showcasing a situation where $v_{prod}$ is higher than $c_{ST}$. This scenario prompts all methods to implement a sidetrack upon any exit from the reservoir, ensuring 100 percent contact for each method. In this geological realization, greedy optimization performs four sidetracks, in contrast to a single sidetrack by both DSDP and RL-Sensor. With 100 percent reservoir contact, the primary factor influencing the overall reward is the number of sidetrack, with DSDP and RL-Sensor each achieving an overall reward that is higher by $3 * c_{ST}$ compared to greedy optimization.

\begin{table}[ht]
\centering
\caption{Scenario 3 - Results for Greedy Optimization, DSDP, and RL-Sensor Robust.}
\begin{tabular}{|c|r|r|r|}
\hline
\textbf{Methods}    & \multicolumn{1}{c|}{\textbf{Rewards}}     & \multicolumn{1}{c|}{\textbf{\begin{tabular}[c]{@{}c@{}}Reservoir\\ Contact(\%)\end{tabular}}} & \multicolumn{1}{c|}{\textbf{\begin{tabular}[c]{@{}c@{}}Operating\\ Cost\end{tabular}}} \\ \hline\hline
\begin{tabular}[c]{@{}c@{}}Greedy\\ \citep{Kullawan2014-2}\end{tabular}            & 104.44                    & 98.26                                  & 9.54                                \\ \hline
\begin{tabular}[c]{@{}c@{}}DSDP\\ \citep{KULLAWAN201890}\end{tabular}              & 107.73                    & 98.26                                  & 6.25                                 \\ \hline
RL-Sensor        & 107.46                    & 97.64                                  & 5.80                                 \\ \hline
\end{tabular}
\label{tab:sce3}
\end{table}

Table \ref{tab:sce3} shows the evaluation rewards, reservoir contact, and operating cost of all three methods for scenario 3. At $v_{prod} = 4$, greedy optimization maximizes immediate reward by performing a sidetrack whenever the well exits the reservoir, resulting in 98.26 percent reservoir contact. However, this policy comes at a significant increase in operating costs. On the other hand, the DSDP reaches the same reservoir contact with fewer sidetracks, resulting in a higher average reward of 107.73 compared to 104.4 for greedy optimization.

The RL-Sensor method achieves a reservoir contact of 97.64 percent but at a lower operating cost than the other methods. Its average reward of 107.46 is only 0.25 lower than the DSDP. However, this is the only scenario where the RL-Sensor method cannot outperform the DSDP, which occurs when the DSDP provides the least additional value over greedy optimization. 

\subsubsection{Discussion}
The results from the second example provide additional evidence supporting our results that the RL-Sensor method outperforms greedy optimization. This example effectively demonstrates the divergence in decision-making between the two methods. Greedy optimization makes sidetrack decisions based solely on the value of $v_{prod}$, whereby if it is lower than $v_{ST}$, greedy optimization always chooses to forego the sidetrack. On the other hand, greedy optimization always chooses to sidetrack if $v_{prod}$ is higher than $v_{ST}$. On the other hand, the sidetrack and overall decision-making policy of the RL-Sensor method consider future values, leading to substantially higher rewards.

Closer investigation to all panels in Figure \ref{fig:ex2single} shows that the RL-Sensor (and DSDP) initiate a downward steering path from the start, aligning closely with the prior information in Figure \ref{fig:priormodel}. In contrast, greedy optimization considers only the immediate information and hence the horizontal (no angle change) initial well trajectory, leading to two reservoir exits. This highlights another key difference in the methods' decision-making policy.

The results from all studied scenarios also indicate that the performance of the RL-Sensor method is comparable to that of the DSDP, which is considered the quasi-optimal solution to the DP. Additionally, the well trajectories in Figure \ref{fig:ex2single} demonstrate the close alignment between the two methods. In several cases, the RL agent even outperforms the DSDP. One possible explanation is that RL does not require discretization, unlike the DSDP, which may lead to better performance. The absence of discretization in RL may enable it to achieve more accurate results by avoiding information loss during the discretization process.

It is worth noting that there is one scenario where the RL-Sensor method does not outperform the DSDP, although the difference is not statistically significant. This scenario is where $v_{prod}$ is higher than $v_{ST}$, leading to a preference for sidetracking over steering if the well exits the reservoir. One plausible explanation is that sidetracking allows to adjust the well trajectory by returning it to the center of the boundaries, thus mitigating the effects of information loss caused by discretization.

In addition to comparable performance, the RL agent substantially reduces long-term computational costs compared to the DSDP. Specifically, after training for approximately 20 minutes, the RL agent can evaluate one geological realization in less than 10 milliseconds. On the other hand, the current discretization setup requires the DSDP to evaluate one geological realization in 15 seconds, 1500 times longer than what RL requires.

\section{Conclusions}\label{conc}
This study introduces and illustrates the application of RL as a flexible, robust, and computationally efficient sequential decision-making tool in two distinct and published geosteering environments. 

The results from the first example indicate that RL-Posterior method suggests decisions that lead to significantly improved value function results compared with greedy optimization, with a 19 to 33 percent increase depending on the scenario. In addition to the RL-Posterior method, we introduce an alternative method called the RL-Sensor method. The RL-Sensor method ignores the posterior updates and relies on the inputs to the Bayesian framework. The RL-Sensor method offers slightly better rewards than the RL-Posterior method while significantly reducing the computational cost. Specifically, the computation time is reduced from 2500 seconds to 500 seconds.

Our results from the second example show that RL provides comparable rewards to the DSDP, which we define as the quasi-optimal solution to the DP. Notably, RL achieves these results with significantly less computational cost. Specifically, the computation time for evaluating one geological realization after training is approximately 10 milliseconds. On the other hand, the DSDP requires 15 second to evaluate one geological realization. These results suggest that RL is a promising alternative to the DSDP for optimizing real-time geosteering decision-making problems where computational efficiency is important.

We also highlight the ease of implementing RL method, which is independent of the environment. The challenge lies in defining the appropriate state space to optimize the reward function for each environment. As shown in our study, certain information may be relevant in one setting but not another. For instance, inclination is included as one of the states in the first environment but not in the second. Overall, the results show the flexibility and potential of RL-based methods in optimizing complex geosteering decision-making problems in various environments.

Our study assumes that the primary source of uncertainty in both environments is the distance to reservoir boundaries and that there are no errors in the sensor readings. In reality, geosteering decisions are made in the face of multiple uncertainties, and sensor readings have different levels of precision. To address this limitation, future studies could explore the ability of RL in providing an efficient and robust geosteering decisions under any number of uncertainties and log readings while dealing with sensor reading errors. This would provide a more realistic evaluation of the performance of RL method and its potential for decision optimization in practical and field case applications.

\section*{Acknowledgements}
This work is part of the Center for Research-based Innovation DigiWells: Digital Well Center for Value Creation, Competitiveness and Minimum Environmental Footprint (NFR SFI project no. 309589, https://DigiWells.no). The center is a cooperation of NORCE Norwegian Research Centre, the University of Stavanger, the Norwegian University of Science and Technology (NTNU), and the University of Bergen. It is funded by Aker BP, ConocoPhillips, Equinor, TotalEnergies, Vår Energi, Wintershall Dea, and the Research Council of Norway.

Sergey Alyaev was supported by the “Distinguish” project during the revision of the manuscript. The project is funded by Aker BP, Equinor, and the Research Council of Norway (RCN PETROMAKS2 project no. 344236).

\section*{Declaration of generative AI and AI-assisted technologies in the writing process}

During the preparation of this work, the author(s) used ChatGPT in order to improve the readability. After using this tool/service, the author(s) reviewed and edited the content as needed and take(s) full responsibility for the content of the publication.


 \bibliographystyle{elsarticle-harv} 
 \bibliography{main}





\end{document}